%% file: main.tex
\documentclass[10pt,journal,compsoc]{IEEEtran}
\ifCLASSOPTIONcompsoc
  \usepackage[nocompress]{cite}
\else
  \usepackage{cite}
\fi
\usepackage{amsmath,amsfonts,amssymb}
\usepackage{algorithmic}
\usepackage{array}
\ifCLASSOPTIONcompsoc
  \usepackage[caption=false,font=footnotesize,labelfont=sf,textfont=sf]{subfig}
\else
  \usepackage[caption=false,font=normalsize,labelfont=sf,textfont=sf]{subfig}
\fi
\usepackage[table]{xcolor}
\usepackage{booktabs}
\usepackage{textcomp}
\usepackage{stfloats}
\usepackage{url}
\usepackage{verbatim}
\usepackage{graphicx}
\usepackage{algorithm}
\usepackage{tabularx}
\usepackage{multirow}
\usepackage{multicol}
\usepackage{pgfplots}
\usepgfplotslibrary{groupplots}
\pgfplotsset{compat=1.18}
\input{preamble}

\def\BibTeX{{\rm B\kern-.05em{\sc i\kern-.025em b}\kern-.08em
    T\kern-.1667em\lower.7ex\hbox{E}\kern-.125emX}}
\usepackage{balance}
\usepackage{ragged2e}
\definecolor{ieeelinkblue}{RGB}{0,102,204}
\usepackage[
    colorlinks=true,
    linkcolor=ieeelinkblue,
    citecolor=ieeelinkblue,
    urlcolor=ieeelinkblue,
    breaklinks=true
]{hyperref}
\begin{document}
\title{Sparse2comm: Towards Robust Cooperative
3D Object Detection}

\author{Lei~Yang, Boqi~Li, Chunmian~Lin, Li~Wang, Ziying~Song, Shaoqing~Xu, Heye~Huang, Haibao~Yu, Chen~Lv
\IEEEcompsocitemizethanks{
\IEEEcompsocthanksitem Lei Yang and Chen Lv are with the School of Mechanical and Aerospace Engineering, Nanyang Technological University, Singapore (e-mail:\{lei.yang,lyuchen\}@ntu.edu.sg).
\IEEEcompsocthanksitem Boqi Li and Chunmian Lin are with the School of Transportation Science and Engineering, Beihang University, Beijing, China, (Email:\{boqili,cmlin\}@buaa.edu.cn).
\IEEEcompsocthanksitem Li Wang is with the School of Mechanical Engineering, Beijing Institute of Technology, Beijing, China. (Email: wangli\_bit@bit.edu.cn.).
\IEEEcompsocthanksitem Ziying Song is with the School of Artificial Intelligence, Yanshan University, China.
\IEEEcompsocthanksitem Shaoqing Xu is with the University of Macau, China. (Email: shaoqing.xu@macau.edu.cn)
\IEEEcompsocthanksitem Heye Huang is with Cho Chun Shik Graduate School of Mobility, Korea Advanced Institute of Science and Technology, Daejeon, Republic of Korea(e-mail: heye.huang@kaist.ac.kr).
\IEEEcompsocthanksitem Haibao Yu is with the University of Hong Kong, China. (Email: yuhaibao94@gmail.com)
}
}

\markboth{Journal of \LaTeX\ Class Files,~Vol.~18, No.~9, September~2020}%
{How to Use the IEEEtran \LaTeX \ Templates}

% Canonical abstract is maintained inline for IEEEtran compsoc compatibility.
\IEEEtitleabstractindextext{%
\begin{abstract}
\justifying
Cooperative perception improves autonomous driving by sharing complementary observations among vehicles and roadside infrastructure for 3D object detection. However, practical deployment is constrained by limited bandwidth and unreliable cooperation, where packet loss, transmission delay, and spatial misalignment jointly degrade the cooperative feature stream. Existing methods often reduce communication cost or compensate for one degradation type, leaving coupled disturbances insufficiently addressed.
To address this problem, we propose Sparse2comm, a bandwidth-efficient and robust cooperative 3D object detection framework that treats unreliable cooperation as progressive restoration over degraded cooperative features. Sparse Feature Encoding first encodes communication as randomly mask-sampled foreground features transmitted by collaborating agents, from which the ego vehicle reconstructs dense semantic representations. This sparse-to-dense mechanism learns to infer missing object-centric content from sparse observations, enabling ultra-low-bandwidth communication and packet-loss recovery within the same representation. On the semantically restored features, Latency-Aware Alignment predicts motion flow to compensate delayed messages, and Self-Calibrating Fusion estimates residual spatial offsets in a self-supervised manner before adaptive cross-agent fusion. Sparse2comm therefore restores semantic completeness, temporal consistency, and spatial alignment in an ordered pipeline.
Extensive experiments on DAIR-V2X, OpenV2V, and V2V4Real show that Sparse2comm maintains competitive clean accuracy and consistently improves robustness under individual and mixed real-world degradations. Compared with the selective feature communication baseline Where2comm, Sparse2comm improves mixed-setting AP@0.5/AP@0.7 by +20.15/+11.79, +12.66/+11.07, and +15.36/+12.61 on the three datasets, respectively. The source code is available at
\url{https://github.com/yanglei18/Sparse2comm}.

\end{abstract}

}
\maketitle

\input{latex/sec/1_intro}
\input{latex/sec/2_relatedworks}
\input{latex/sec/3_method}
\input{latex/sec/4_exps}
\input{latex/sec/5_conclusion}

%%%%%%%%% REFERENCES
{\small
\bibliographystyle{IEEEtran}
\bibliography{main}
}

\end{document}

%% file: preamble.tex
\makeatletter
\ifCLASSOPTIONcompsoc
\ifCLASSOPTIONconference\else
\long\def\@makecaption#1#2{%
\ifx\@captype\@IEEEtablestring%
\footnotesize\bgroup
\noindent\parbox[t]{\hsize}{\normalfont\sffamily\footnotesize
\@IEEEtabletopskipstrut{#1:}~#2}\par
\addvspace{0.5\baselineskip}\egroup%
\@IEEEtablecaptionsepspace
\else
\@IEEEfigurecaptionsepspace
\setbox\@tempboxa\hbox{\normalfont\sffamily\footnotesize {#1.}\nobreakspace #2}%
\ifdim \wd\@tempboxa >\hsize%
\setbox\@tempboxa\hbox{\normalfont\sffamily\footnotesize {#1.}\nobreakspace}%
\parbox[t]{\hsize}{\normalfont\sffamily\footnotesize \noindent\unhbox\@tempboxa#2}%
\else%
\hbox to\hsize{\normalfont\sffamily\footnotesize\box\@tempboxa\hfil}%
\fi\fi}
\fi
\fi
\makeatother

%% file: latex/sec/1_intro.tex
\section{Introduction}
\input{latex/fig/teaser}

Cooperative perception has emerged as a key paradigm for enhancing autonomous driving safety, particularly in complex urban scenes where occlusion, restricted fields of view, and limited sensing range constrain ego-vehicle perception. By enabling an ego vehicle to exchange complementary observations with collaborating agents, including nearby vehicles and roadside units (RSUs), cooperative perception provides a more complete understanding of the surrounding environment and improves the reliability of 3D object detection in safety-critical scenarios~\cite{yu2022dair,yu2025end,yu2023flow,yu2023v2x,liu2025toward,gao2024survey}. However, moving from this promise to real-world deployment requires cooperative features to be both bandwidth-efficient and robust to unreliable communication and calibration. Current C-V2X links typically provide less than 10 Mbps~\cite{clancy2024wireless}, making it impractical to transmit raw point clouds or dense feature maps at high frame rates. Although feature compression and object-level communication reduce data volume~\cite{hu2022where2comm,liu2024v2x,yang2023how2comm,hu2026pragmatic}, excessive compression may remove discriminative semantic cues and degrade detection reliability. Beyond this bandwidth constraint, cooperative messages are further affected by three major robustness degradations. \textbf{Packet loss} corrupts transmitted features under interference, congestion, and unstable wireless links~\cite{gao2024survey}; even a 5\% packet loss can reduce cooperative perception accuracy by nearly 37\%~\cite{tang2025rocooper}. \textbf{Transmission delay} introduces temporal asynchrony between collaborator-side and ego-vehicle observations, especially for moving objects whose locations change rapidly across frames~\cite{song2025traf,lei2022latency,wei2023asynchrony}. \textbf{Spatial misalignment} arises from GNSS multipath effects, sensor drift, and calibration errors, causing collaborator features to be projected to inaccurate positions in the ego-vehicle coordinate frame~\cite{song2023cooperative}. These factors often co-occur in complex scenes and jointly degrade the cooperative feature stream before fusion, making received messages sparse, incomplete, temporally stale, and spatially biased.

Existing studies have made important progress on individual aspects of this problem. For bandwidth constraints, selective communication methods transmit only informative regions or compact object representations. For example, Where2comm~\cite{hu2022where2comm} uses spatial confidence maps to share critical feature regions, while V2X-PC~\cite{liu2024v2x} clusters points into compact packages to preserve object structure. For packet loss, recent works explore feature recovery and history-aided reconstruction. V2X-INCOP~\cite{ren2024interruption}, for instance, leverages temporal redundancy to restore corrupted or missing cooperative messages. For temporal asynchrony, trajectory or flow-based methods compensate delayed messages using historical context~\cite{song2025traf,lei2022latency,wei2023asynchrony}. For spatial misalignment and noisy cooperation, alignment-based and robust communication approaches refine relative poses, aggregate motion-aware features, or perform uncertainty-aware asynchronous fusion~\cite{10160546,hong2024mrcnet,xu2025codyntrust}. However, most existing methods are designed around either communication reduction or a specific degradation type. In real deployments, these factors are coupled: aggressive compression can amplify the impact of packet loss, delayed messages can worsen spatial inconsistency for moving objects, and inaccurate poses can make otherwise informative features harmful during fusion. Therefore, robust cooperative perception should not treat communication efficiency, packet-loss recovery, temporal alignment, and spatial calibration as isolated remedies. It requires an ordered restoration process that progressively recovers semantic completeness, temporal consistency, and spatial alignment from the same degraded feature stream.

To address this gap, we propose Sparse2comm, a bandwidth-efficient and robust cooperative 3D object detection framework that treats unreliable cooperation as progressive restoration over degraded cooperative features. In this formulation, communication loss, temporal asynchrony, and spatial misalignment are not treated as independent disturbances. Instead, the received cooperative message is regarded as a degraded observation of an underlying cooperative feature state: it is sparse due to bandwidth constraints, incomplete due to packet loss, temporally stale due to transmission delay, and spatially biased due to localization or calibration errors. The objective is to recover a compact, semantically complete, temporally synchronized, and spatially calibrated cooperative representation before final detection.
To instantiate this restoration paradigm, Sparse2comm organizes Sparse Feature Encoding, Latency-Aware Alignment, and Self-Calibrating Fusion into a dependent sparse-to-aligned restoration pipeline. Sparse Feature Encoding (SFE) first encodes communication as randomly mask-sampled foreground features transmitted by collaborating agents, from which the ego vehicle reconstructs dense semantic representations. By training the ego vehicle to infer missing object-centric content from sparse foreground observations, SFE turns bandwidth reduction and packet-loss recovery into the same sparse-to-dense restoration problem. Given the semantically restored features from SFE, Latency-Aware Alignment (LAA) predicts motion flow from temporal collaborator features to compensate for transmission delay and synchronize cooperative features with the ego-vehicle timestamp. This second stage is placed after SFE because temporal alignment is more reliable when object-centric semantic content has been partially recovered. After temporal synchronization, Self-Calibrating Fusion (SCF) estimates residual spatial offsets in a self-supervised manner and refines adaptive cross-agent fusion. Placing SCF after LAA helps separate pose-induced offsets from temporal staleness, allowing final fusion to operate on both temporally synchronized and spatially calibrated features. These stages are not independent add-ons: SFE restores semantic completeness from sparse and lossy observations, LAA restores temporal consistency on recovered features, and SCF restores spatial alignment after temporal synchronization.

Extensive experiments on DAIR-V2X~\cite{yu2022dair}, OpenV2V~\cite{xu2022opv2v}, and V2V4Real~\cite{xu2023v2v4real} demonstrate that Sparse2comm maintains competitive clean accuracy and generalizes across vehicle--infrastructure, simulated vehicle--vehicle, and real-world vehicle--vehicle cooperative perception scenarios. Under the mixed robustness setting with packet loss, transmission delay, and spatial misalignment, Sparse2comm improves over the strongest prior methods by +15.75/+9.37, +7.13/+4.11, and +11.11/+7.34 AP@0.5/AP@0.7 on the three datasets, respectively. Compared with the representative selective communication baseline Where2comm~\cite{hu2022where2comm}, the corresponding mixed-setting gains are +20.15/+11.79, +12.66/+11.07, and +15.36/+12.61 AP@0.5/AP@0.7, while Sparse2comm uses only 1.0\% relative feature bandwidth. These results show that Sparse2comm provides a strong trade-off among detection accuracy, robustness, and communication efficiency.

The main contributions are summarized as follows:
\begin{itemize}
\item We formulate bandwidth-limited and unreliable cooperative 3D object detection as progressive restoration over degraded cooperative features across vehicle--infrastructure and vehicle--vehicle settings, enabling robust cooperative perception with only 1.0\% relative feature bandwidth compared with dense feature-level communication.
\item We design a dependent SFE--LAA--SCF restoration pipeline: SFE transmits randomly mask-sampled foreground features and reconstructs dense semantic representations for ultra-low-bandwidth communication and packet-loss recovery, LAA restores temporal consistency, and SCF restores spatial alignment before adaptive fusion.
\item We provide comprehensive evaluations on DAIR-V2X~\cite{yu2022dair}, OpenV2V~\cite{xu2022opv2v}, and V2V4Real~\cite{xu2023v2v4real} under clean and robustness-critical settings, showing that Sparse2comm generalizes across diverse cooperative perception scenarios. Under mixed degradations, Sparse2comm improves over Where2comm~\cite{hu2022where2comm} by \textbf{+20.15/+11.79}, \textbf{+12.66/+11.07}, and \textbf{+15.36/+12.61} AP@0.5/AP@0.7 on the three datasets, respectively, while maintaining a compact \textbf{1.0\%} relative feature bandwidth.

\end{itemize}

%% file: latex/fig/teaser.tex
\begin{figure*}[h!t]
\centering
\setlength{\abovecaptionskip}{2pt}
\includegraphics[width=18.0cm]{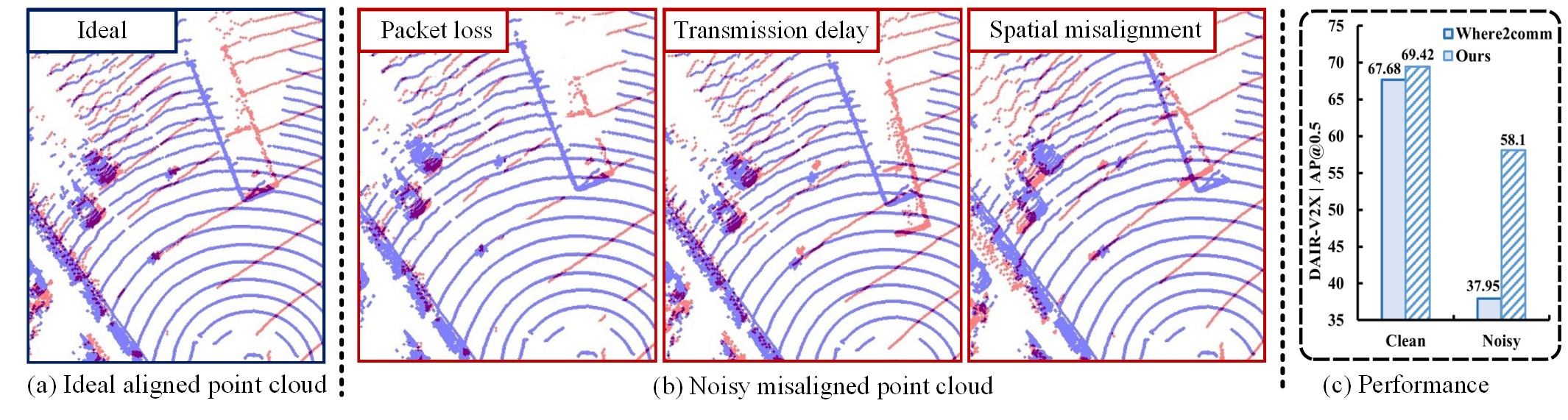}
\caption{\textbf{Illustration of real-world challenges in cooperative perception.} 
(a) Ideal aligned point cloud: ideal scenario with accurately synchronized and aligned data between ego and collaborating sensors. 
(b) Noisy misaligned point cloud: real-world distortions caused by (left) packet loss, (middle) transmission delay, and (right) localization errors, which jointly lead to incomplete or misaligned spatial representations. 
(c) Performance degradation: standard methods suffer significant accuracy drops under these noisy conditions, whereas our proposed Sparse2comm maintains robust cooperative perception under bandwidth, latency, and spatial uncertainties.}
\label{fig:teaser}
\vspace{-0.3cm}
\end{figure*}

%% file: latex/sec/2_relatedworks.tex
\section{Related Works}

\subsection{Ego-centric 3D Object Detection}
Accurate 3D object detection is a fundamental capability for autonomous driving. Existing ego-centric detectors are commonly developed around sensing modality and spatial representation. Vision-based methods~\cite{wang2021fcos3d,yang2023lite,yang2023mix} estimate 3D object geometry from monocular or multi-view images, offering a low-cost perception solution but remaining sensitive to depth ambiguity, illumination changes, and adverse weather. Recent camera-based BEV frameworks~\cite{li2024bevformer,li2023bevdepth,yang2023bevheight,yang2025bevheight++} lift image features into a unified bird's-eye-view space, which makes cross-view spatial reasoning more explicit. LiDAR-based detectors~\cite{zhou2018voxelnet,yan2018second,lang2019pointpillars}, represented by VoxelNet~\cite{zhou2018voxelnet}, SECOND~\cite{yan2018second}, and PointPillars~\cite{lang2019pointpillars}, exploit accurate geometric measurements through voxel or pillar representations. Multi-modal detectors~\cite{liu2022bevfusion,song2023graphalign,song2024graphbev} further combine LiDAR and camera cues to integrate appearance and geometry.

Although these ego-centric detectors substantially improve local perception, they still operate under a single-platform sensing assumption. Objects hidden by occlusion, distant traffic participants, and complex intersection layouts can remain unobservable from one viewpoint, regardless of the local detector architecture. These limitations motivate cooperative perception, where vehicles, roadside units, and other collaborating agents exchange complementary observations to extend the effective perception range and reduce blind spots. This shift also introduces communication and reliability constraints that are absent in purely ego-centric detection.

\subsection{Cooperative Perception}
Cooperative perception has developed rapidly with the emergence of V2X benchmarks, including simulated datasets such as V2X-Sim~\cite{li2022v2xsim} and OPV2V~\cite{xu2022opv2v}, system-level simulation platforms for collaborative autonomous driving~\cite{liu2025toward}, real-world vehicle--infrastructure datasets such as DAIR-V2X~\cite{yu2022dair} and V2X-Seq~\cite{yu2023v2x}, and recent real-world V2V/V2X or multi-modal datasets~\cite{xu2023v2v4real,xiang2024v2x,yang2024v2x}. These benchmarks have shifted the field from isolated ego perception to collaborative scene understanding, and they also expose practical constraints caused by limited bandwidth, unreliable transmission, and imperfect localization.

Existing cooperative methods are commonly organized into early, late, and intermediate fusion. Early fusion preserves rich raw observations but incurs heavy communication, whereas late fusion is efficient but discards dense spatial evidence before cross-agent reasoning. Intermediate fusion exchanges feature representations and has become a practical balance between accuracy and bandwidth. Representative methods include feature-level fusion and message-passing approaches~\cite{chen2019f,wang2020v2vnet,liu2020when2com}, distillation or transformer-based cooperative fusion~\cite{mehr2019disconet,xu2022v2x,xu2025v2xvitv2,xu2023cobevt}, and communication-aware variants~\cite{hu2022where2comm,yang2023what2comm,yang2023how2comm,zhang2024ermvp,xu2025cosdh,hu2026pragmatic}. Most of these methods are evaluated under relatively reliable feature exchange, where received messages are assumed to be usable after transmission.

Communication efficiency is a key requirement for practical cooperative perception over bandwidth-limited V2X links. Selective, pragmatic, and object-aware methods reduce transmitted data by sharing informative regions, compact object messages, or hybrid intermediate-late representations~\cite{hu2022where2comm,liu2024v2x,ding2025point,zhang2024ermvp,xu2025cosdh,hu2026pragmatic}. These methods show that sparse communication can greatly reduce transmission cost. However, sparsification and packet loss both remove feature content from the cooperative stream, so message compactness and message recoverability are tightly coupled. This coupling suggests that communication-efficient representations should also be recoverable under missing observations.

Robustness to communication and calibration errors is another important direction. Prior works have studied lossy or interrupted communication~\cite{li2023learning,ren2024interruption,tang2025rocooper}, transmission delay and temporal asynchrony~\cite{lei2022latency,wei2023asynchrony,song2025traf}, and spatial misalignment caused by localization or pose errors~\cite{10160546,song2023cooperative}. More recent robust cooperative perception methods further explore motion-aware feature aggregation and uncertainty-aware asynchronous fusion~\cite{hong2024mrcnet,xu2025codyntrust}. Most of these methods target a specific degradation type. In practice, these disturbances often occur together, so robustness needs to be examined under packet loss, delay, spatial misalignment, and their coupled setting rather than as isolated cases.

%% file: latex/sec/3_method.tex
\section{Sparse2comm Framework}

\subsection{Problem Definition}
\label{sec:problem_definition}
Cooperative perception aims to improve ego-vehicle 3D object detection by fusing complementary features from collaborating agents, which can be nearby vehicles or roadside units (RSUs), under a communication budget. Formally, the cooperative detection process can be written as
\begin{equation}
\hat{B}_{\text{ego}} =
{\Psi}_{\text{Coop.}}^{\omega}
\left(F_v^{t_v}, \left\{F_i^{t_i}\right\}_{i=1}^{N}\right),
\quad
\text{s.t.} \quad
\sum_{i=1}^{N}{\left\|F_i^{t_i}\right\|} \le C ,
\end{equation}
where ${\Psi}_{\text{Coop.}}^{\omega}$ denotes the cooperative perception model parameterized by $\omega$, and $\hat{B}_{\text{ego}}$ is the set of predicted 3D bounding boxes for the ego vehicle. $F_v^{t_v}$ denotes the ego-vehicle feature at timestamp $t_v$, while $F_i^{t_i}$ denotes the feature transmitted by the $i$-th collaborating agent at timestamp $t_i$. $N$ is the number of collaborating agents, and $C$ is the communication bandwidth constraint that limits the aggregate transmitted feature volume. The transmission delay between the collaborating agent and the ego vehicle is denoted as $\tau = t_v - t_i$. In real deployments, the received cooperative features are further degraded by packet loss, temporal asynchrony, and spatial misalignment. Sparse2comm therefore aims to recover a compact, semantically complete, temporally synchronized, and spatially calibrated cooperative representation before final detection.

\input{latex/fig/framework}

\subsection{Overall Architecture}
Sparse2comm follows a progressive restoration pipeline that sequentially restores semantic completeness, temporal consistency, and spatial alignment before cooperative fusion.
The overall pipeline in Fig.~\ref{fig:framework} starts from collaborator-side feature extraction. The collaborator encoder, built upon a PFE~\cite{lang2019pointpillars} and a ResNet~\cite{he2016deep} backbone, extracts a sequence of BEV feature maps $\left\{F_i^{t_i-3}, \ldots, F_i^{t_i}\right\}$ from collaborator-side point clouds $\left\{P_i^{t_i-3}, \ldots, P_i^{t_i}\right\}$, where $F_i^{t_i} \in \mathbb{R}^{X_i \times Y_i \times C_i}$. Here, $X_i$, $Y_i$, and $C_i$ denote the height, width, and channel dimensions of the collaborator BEV feature, respectively. Based on these features, the RPN predicts ROI boxes $B_i^{t_i} = \left\{b_o^{t_i}\right\}_{o=1}^{O}$, where $O$ is the number of detected objects. Guided by these ROI boxes, an object-aware filter~\cite{hu2022where2comm} removes redundant background responses and produces object-foreground BEV features $\left\{\mathcal{F}_i^{t_i-3}, \ldots, \mathcal{F}_i^{t_i}\right\}$. The ego encoder, which adopts the same PFE-ResNet backbone, extracts ego-vehicle BEV features $F_v^{t_v} \in \mathbb{R}^{X_v \times Y_v \times C_v}$ from the ego-vehicle point cloud $P_v^{t_v}$.

The object-foreground feature $\mathcal{F}_i^{t_i}$ is first processed by Sparse Feature Encoding. Instead of transmitting dense feature maps, the collaborating agent sends only a masked sparse feature $\widetilde{\mathcal{F}}_i^{t_i}$, and the ego vehicle reconstructs it into $\hat{\mathcal{F}}_i^{t_i} \in \mathbb{R}^{X_i \times Y_i \times C_i}$. Latency-Aware Alignment uses temporal collaborator features to estimate motion flow and warps the recovered feature to the ego-vehicle timestamp, producing $\hat{\mathcal{F}}_i^{t_v}$. The temporally aligned collaborator feature is transformed into the ego-vehicle coordinate frame and refined by Self-Calibrating Fusion, which estimates residual spatial offsets and fuses the calibrated collaborator feature with $F_v^{t_v}$. The resulting fused feature $Z_v^{t_v} \in \mathbb{R}^{X_v \times Y_v \times C_v}$ is finally decoded by a detection head to predict $\hat{B}_{\text{ego}}$.

\subsection{Sparse Feature Encoding}
Sparse Feature Encoding serves as the first restoration stage and provides the core mechanism for both low-bandwidth communication and packet-loss recovery. Different from selective communication methods that transmit selected foreground responses~\cite{hu2022where2comm}, SFE encodes communication as randomly mask-sampled foreground features. The collaborating agent sends only a small subset of object-centric BEV patches, and the ego vehicle reconstructs dense semantic representations from these sparse observations. In this view, bandwidth reduction is not a post-hoc compression step. It becomes a sparse-to-dense reconstruction problem, where intentionally omitted patches and accidentally lost packets are handled as missing foreground observations under the same reconstruction mechanism.

Fig.~\ref{fig:motivation_sparse_feature_encoding} motivates this sparse communication design: the collaborating agent applies random mask-based sampling to the object-foreground BEV feature $\mathcal{F}_i^{t_i}$ and obtains an ultra-sparse transmitted representation $\widetilde{\mathcal{F}}_i^{t_i}$. With a masking ratio
\begin{equation}
r = 1 - \frac{\left\|\widetilde{\mathcal{F}}_i^{t_i}\right\|}{\left\|\mathcal{F}_i^{t_i}\right\|},
\end{equation}
only about 1\% of conventional dense feature-level communication needs to be transmitted. The ego vehicle then reconstructs dense semantic representations $\hat{\mathcal{F}}_i^{t_i}$ from the received sparse features. Because the reconstruction objective is trained to infer missing object-foreground content, the same mechanism recovers both intentionally omitted patches and accidentally lost packets, providing robustness under unstable communication.
\input{latex/fig/motivation_sparse_feature_encoding}

Sparse Feature Encoding is implemented with two coupled stages, detailed in Fig.~\ref{fig:sparse_feature_encoding}: Mask Sampling on the collaborator side and Feature Reconstruction on the ego-vehicle side.

\subsubsection{\textbf{Mask Sampling}}
Mask Sampling minimizes transmission cost while preserving essential object-level cues. It takes the object-foreground BEV feature $\mathcal{F}_i^{t_i}$ as input and outputs an ultra-sparse feature map $\widetilde{\mathcal{F}}_i^{t_i}$ that retains only partial foreground information. Different from global random masking, the proposed strategy restricts sampling to object regions and independently samples visible patches for each object instance. The random object-wise sampling is deliberately used as a missing-pattern generator: during training, the reconstruction network observes diverse visible patch subsets and learns to infer object-centric content from incomplete evidence rather than relying on fixed transmitted positions. This object-wise design reduces the risk of removing an entire object and introduces instance-level sparsity variation that improves robustness to communication uncertainty.
\input{latex/fig/sparse_feature_encoding}

Formally, let $\mathcal{F}_i^{t_i} \in \mathbb{R}^{X_i \times Y_i \times C_i}$ denote the object-foreground BEV feature after background removal. We divide $\mathcal{F}_i^{t_i}$ into $N_{\mathrm{fg}}$ non-overlapping patches:
\begin{equation}
\mathcal{P}_i^{t_i} = \{ f_1, f_2, \ldots, f_{N_{\mathrm{fg}}} \}, \quad
f_j \in \mathbb{R}^{p \times p \times C_i}.
\end{equation}
Each patch $f_j$ is associated with an object according to ROI box $b_o^{t_i}$, forming an object-wise partition:
\begin{equation}
\mathcal{J}_o = \{ j \mid f_j \in b_o^{t_i} \}, \quad
N_o = |\mathcal{J}_o|, \quad o=1,\ldots,O.
\end{equation}
For each object $o$, we sample a random masking ratio:
\begin{equation}
r_o \sim \mathcal{U}(r_{\min}, r_{\max}), \quad
0 < r_{\min} \le r_{\max} < 1,
\end{equation}
and retain $K_o$ visible patches:
\begin{equation}
K_o = \left\lfloor (1-r_o) \cdot N_o \right\rfloor .
\end{equation}
The visible patch subset for object $o$ is sampled as
\begin{equation}
V_o = \operatorname{Sample}(\mathcal{J}_o, K_o).
\end{equation}
The full visible set is obtained by aggregating all object-wise visible subsets:
\begin{equation}
V = \bigcup_{o=1}^{O} V_o.
\end{equation}
To unify intentional sparsification and packet loss, we distinguish the sampled visible set from the actually received visible set. During transmission, a subset of the sampled patches may be dropped due to packet loss, denoted as $V_{\mathrm{drop}} \subseteq V$. The ego vehicle therefore observes
\begin{equation}
V_{\mathrm{rec}} = V \setminus V_{\mathrm{drop}} .
\end{equation}
The received ultra-sparse BEV feature map is then constructed as
\begin{equation}
\widetilde{\mathcal{F}}_i^{t_i} =
\{(f_j, \mathrm{pos}_j) \mid j \in V_{\mathrm{rec}}\}.
\end{equation}
Accordingly, the missing patch set to be reconstructed is
\begin{equation}
\mathcal{M}
=
\left(\{1,\ldots,N_{\mathrm{fg}}\} \setminus V\right)
\cup
V_{\mathrm{drop}},
\end{equation}
which contains both intentionally omitted patches and packet-lost patches. This formulation makes packet loss a natural extension of the same missing-content reconstruction problem used for bandwidth reduction. This design retains object-level evidence under an extremely small communication budget while exposing the reconstruction network to diverse sparsity patterns during training.

\subsubsection{\textbf{Feature Reconstruction}}
Feature Reconstruction is the semantic restoration component of SFE. It restores incomplete object-foreground features caused by intentional mask sampling and packet loss during transmission. Since both effects remove foreground patches from the received observation, they can be recovered by the same sparse-to-dense decoder. It adopts an encoder-decoder architecture that learns a reconstruction prior over object-centric BEV features.

The encoder consists of five ViT~\cite{vaswani2017attention} blocks. Each received foreground patch is projected by a linear layer, augmented with positional encoding, and processed by the encoder to produce a high-dimensional embedding. Only patches indexed by $V_{\mathrm{rec}}$ are forwarded to the encoder, while masked or packet-lost patches are omitted, which keeps the computation compatible with sparse communication.

The decoder consists of two ViT~\cite{vaswani2017attention} blocks followed by a linear layer. Its input contains the encoder embeddings of visible patches and placeholder embeddings for the missing patch set $\mathcal{M}$, where each missing patch is represented by a shared learnable mask token. After adding positional encodings, the decoder reconstructs the full set of object-foreground patches and rearranges them into the recovered BEV feature map $\hat{\mathcal{F}}_i^{t_i}$.

The reconstruction process is optimized by minimizing the mean squared error between the recovered feature $\hat{\mathcal{F}}_i^{t_i}$ and the original object-foreground feature $\mathcal{F}_i^{t_i}$:
\begin{equation}
\mathcal{L}_{\text{REC}}
({\hat{\mathcal{F}}}_i^{t_i}, \mathcal{F}_i^{t_i})
=
\frac{1}{X_iY_i}
\sum_{m,n}^{X_iY_i}
\left[
{\hat{\mathcal{F}}}_i^{t_i}(m,n)
-
\mathcal{F}_i^{t_i}(m,n)
\right]^2 .
\end{equation}
In this way, Sparse Feature Encoding restores semantic completeness before any temporal or spatial correction is applied, providing a more reliable feature basis for the subsequent alignment stages.

\subsection{Latency-Aware Alignment}
After SFE restores semantic completeness, the recovered collaborator feature can still be temporally stale because it is generated at the collaborator timestamp rather than the ego timestamp. For moving objects, this temporal asynchrony appears as spatial displacement in the BEV plane and can degrade feature fusion even when the transmitted content has been reconstructed. Latency-Aware Alignment is therefore placed after SFE: it estimates motion flow from recent collaborator features and warps the recovered feature to the ego-vehicle timestamp, restoring temporal consistency before spatial calibration. The module decomposes this process into flow generation and temporal warping, as summarized in Fig.~\ref{fig:latency_aware_alignment}.

\subsubsection{\textbf{Flow Generator}}
The Flow Generator predicts pixel-wise motion $\mathcal{V}_i^{t_i}$ from a temporal sequence of collaborator BEV features. Given $\{\mathcal{F}_i^{t_i-3}, \ldots, \mathcal{F}_i^{t_i}\}$, we concatenate the features along the channel dimension:
\begin{equation}
\mathbf{X}_{\text{flow}} =
\left[\mathcal{F}_i^{t_i-3}, \ldots, \mathcal{F}_i^{t_i}\right].
\end{equation}
The Flow Generator follows an encoder-decoder architecture with three downsampling stages, three upsampling stages, and a final convolutional layer. Each downsampling unit contains a convolutional layer, Batch Normalization (BN)~\cite{ioffe2015batch}, and ReLU~\cite{nair2010rectified}; each upsampling unit uses a deconvolution layer, BN, and ReLU. The predicted motion flow is
\begin{equation}
\mathcal{V}_i^{t_i} =
f_{\text{flow}}(\mathbf{X}_{\text{flow}}), \quad
\mathcal{V}_i^{t_i} \in \mathbb{R}^{X_i \times Y_i \times 2},
\end{equation}
where $\mathcal{V}_i^{t_i}(m,n)$ denotes the BEV motion vector at position $(m,n)$.
\input{latex/fig/latency_aware_alignment}

\subsubsection{\textbf{Temporal Alignment}}
Given the predicted flow, a bilinear warping function $f_{\text{warp}}(\cdot)$ aligns the collaborator feature to the ego-vehicle timestamp:
\begin{equation}
\hat{\mathcal{F}}_i^{t_v}
=
f_{\text{warp}}
\left(
\mathcal{F}^{t_i},
\mathcal{V}_i^{t_i} \cdot \tau
\right),
\quad
\mathcal{F}^{t_i} =
\begin{cases}
\mathcal{F}_i^{t_i}, & \text{if training}, \\
\hat{\mathcal{F}}_i^{t_i}, & \text{if inference}.
\end{cases}
\end{equation}
During training, the clean object-foreground feature $\mathcal{F}_i^{t_i}$ provides stable supervision for learning motion flow. During inference, the reconstructed feature $\hat{\mathcal{F}}_i^{t_i}$ is warped because only sparse transmitted features are available at the ego vehicle. The flow predictor is supervised by a cosine similarity loss over future collaborator frames:
\begin{equation}
\mathcal{L}_{\text{FLOW}} =
\sum_{t \in \{t_i+\Delta t,\ldots,t_i+n\Delta t\}}
\left(
1 -
\frac{
\mathcal{F}_i^{t} \odot \hat{\mathcal{F}}_i^{t}
}{
\left\|\hat{\mathcal{F}}_i^{t}\right\|_2
\cdot
\left\|\mathcal{F}_i^{t}\right\|_2
}
\right),
\end{equation}
where $\Delta t$ denotes the sampling interval and $n$ is the number of future frames used for temporal supervision.

\subsection{Self-Calibrating Fusion}
After temporal consistency has been restored, the remaining feature offset mainly comes from imperfect relative pose estimation, GNSS multipath effects, sensor drift, or calibration errors. Separating temporal alignment from spatial calibration is important because both delay-induced motion and pose-induced bias appear as spatial displacement in the BEV plane. Let
\begin{equation}
\hat{\mathcal{F}}_v^{t_v}
=
\mathbf{T}_i^v
\left(
\hat{\mathcal{F}}_i^{t_v}
\right)
\end{equation}
denote the temporally aligned collaborator feature transformed into the ego-vehicle coordinate frame by the relative pose $\mathbf{T}_i^v$. If $\mathbf{T}_i^v$ is inaccurate, $\hat{\mathcal{F}}_v^{t_v}$ becomes spatially misaligned with the ego-vehicle feature $F_v^{t_v}$. Self-Calibrating Fusion restores spatial alignment by learning feature-space residual offset calibration in a self-supervised manner before adaptive feature fusion. In contrast to directly correcting the global pose, SCF estimates local residual offsets between cooperative features, which makes it suitable for compensating fine-grained feature misalignment after temporal synchronization. Its two stages, Self-Calibrating Alignment and Adaptive Feature Fusion, are illustrated in Fig.~\ref{fig:self_calibrating_fusion}.
\input{latex/fig/self_calibrating_fusion}

\subsubsection{\textbf{Self-Calibrating Alignment}}
Self-Calibrating Alignment estimates pixel-wise residual offsets between the source collaborator feature fed to the calibration module and the ego-vehicle feature. To avoid mixing the training and inference inputs, we denote this source feature as $\bar{\mathcal{F}}_v^{t_v}$: during self-supervised training, it is the noisy transformed feature defined below; during inference, it is the collaborator feature transformed by the available relative pose. The two feature maps are first concatenated along the channel dimension:
\begin{equation}
\mathbf{X}_{\text{sca}}
=
\left[
\bar{\mathcal{F}}_v^{t_v},
F_v^{t_v}
\right].
\end{equation}
The concatenated feature is processed by a residual block and a deformable convolution layer to predict the offset field:
\begin{equation}
\Delta_{i \to v}
=
f_{\text{DCN}}
\left(
f_{\text{residual}}(\mathbf{X}_{\text{sca}})
\right),
\end{equation}
where $\Delta_{i \to v}$ represents the pixel-wise residual displacement from the collaborator feature to the ego-vehicle feature. The offset field is then applied to the transformed collaborator feature through bilinear sampling:
\begin{equation}
H_v^{t_v}(p_k)
=
\bar{\mathcal{F}}_v^{t_v}
\left(
p_k + \Delta_{i \to v}(p_k)
\right),
\quad
0 \le k \le X_vY_v,
\end{equation}
where $p_k \in \mathbb{R}^{2}$ denotes the BEV pixel coordinate and $H_v^{t_v}$ is the spatially calibrated collaborator feature.

To train this residual calibration without manual offset annotations, we introduce a self-supervised perturbation strategy. During training, the clean relative pose $\mathbf{T}_i^v$ is perturbed by a synthetic global transformation $\mathbf{T}_{\text{noisy}}$ to simulate localization and calibration errors. The noisy transformed feature is
\begin{equation}
\hat{\mathcal{F}}_{v,\text{noisy}}^{t_v}
=
\mathbf{T}_{\text{noisy}}
\mathbf{T}_i^v
\left(
\hat{\mathcal{F}}_i^{t_v}
\right),
\end{equation}
During self-supervised training, we set $\bar{\mathcal{F}}_v^{t_v}=\hat{\mathcal{F}}_{v,\text{noisy}}^{t_v}$ in $\mathbf{X}_{\text{sca}}$, while the clean transformed feature $\hat{\mathcal{F}}_v^{t_v}$ serves as the reconstruction target. This perturb-and-recover objective teaches the offset predictor to remove residual spatial bias without requiring manually annotated calibration errors. The alignment loss is defined as
\begin{equation}
\mathcal{L}_{\text{SCA}}
=
\frac{1}{X_vY_v}
\sum_{m,n}^{X_vY_v}
\left[
H_v^{t_v}(m,n)
-
\hat{\mathcal{F}}_v^{t_v}(m,n)
\right]^2 .
\end{equation}
At inference time, no synthetic perturbation is added, and $\bar{\mathcal{F}}_v^{t_v}$ is the transformed collaborator feature obtained from the deployed relative pose. When the deployed relative pose contains residual error, the learned offset field automatically compensates for spatial deviations and improves geometric consistency before fusion.

\subsubsection{\textbf{Adaptive Feature Fusion}}
After semantic restoration, temporal alignment, and spatial calibration, the calibrated collaborator feature $H_v^{t_v}$ and the ego-vehicle feature $F_v^{t_v}$ are fused through a residual channel-attention module. We first concatenate the two inputs:
\begin{equation}
\mathbf{X}_{\text{fuse}}
=
\left[
H_v^{t_v},
F_v^{t_v}
\right].
\end{equation}
The concatenated feature is refined by a residual block:
\begin{equation}
\begin{aligned}
\mathbf{Y}
&=
\operatorname{ReLU}
\left(
\operatorname{BN}
\left(
W_1\mathbf{X}_{\text{fuse}}
\right)
\right),\\
F
&=
f_{\text{residual}}(\mathbf{X}_{\text{fuse}})\\
&=
\operatorname{ReLU}
\left(
\operatorname{BN}
\left(
W_2\mathbf{Y}
\right)
+ \mathbf{X}_{\text{fuse}}
\right).
\end{aligned}
\label{eq:residual}
\end{equation}
where $\mathbf{Y}$ is the intermediate residual feature, and $W_1$ and $W_2$ are $3{\times}3$ convolution kernels. A squeeze-and-excitation attention layer then reweights feature channels:
\begin{equation}
f_{\text{se}}(F)
=
\sigma
\left(
W f_{\text{avg}}(F)
\right)
\odot F,
\label{eq:se}
\end{equation}
where $f_{\text{avg}}(\cdot)$ denotes global average pooling, $W$ is a learnable weight matrix, $\sigma(\cdot)$ is the sigmoid function, and $\odot$ denotes channel-wise multiplication. The final fused BEV representation is
\begin{equation}
Z_v^{t_v}
=
f_{\text{se}}(F).
\end{equation}
This adaptive fusion stage integrates complementary cross-agent information after the cooperative feature stream has been made semantically complete, temporally synchronized, and spatially calibrated.

\subsection{Detection Head and Loss Function}
\subsubsection{\textbf{Detection Head}}
Following PointPillars~\cite{lang2019pointpillars}, Sparse2comm adopts a single-stage detection head with three branches: classification for object categories, regression for 3D bounding box parameters $(x, y, z, l, w, h, \theta)$, and direction classification for orientation disambiguation. This lightweight head decodes the fused BEV feature $Z_v^{t_v}$ into final 3D bounding boxes.

\subsubsection{\textbf{Overall Loss Function}}
The overall training objective combines the detection loss with auxiliary supervision from the three restoration stages:
\begin{equation}
\mathcal{L}_{\text{total}}
=
\mathcal{L}_{\text{det}}
+
\lambda_1 \mathcal{L}_{\text{REC}}
+
\lambda_2 \mathcal{L}_{\text{FLOW}}
+
\lambda_3 \mathcal{L}_{\text{SCA}},
\end{equation}
where $\lambda_1$, $\lambda_2$, and $\lambda_3$ are balancing weights. $\mathcal{L}_{\text{det}}$ is the 3D detection loss, $\mathcal{L}_{\text{REC}}$ supervises semantic reconstruction in Sparse Feature Encoding, $\mathcal{L}_{\text{FLOW}}$ supervises temporal motion prediction in Latency-Aware Alignment, and $\mathcal{L}_{\text{SCA}}$ supervises residual spatial calibration in Self-Calibrating Fusion.

%% file: latex/fig/framework.tex
\begin{figure*}[h!t]
\centering
\includegraphics[width=1.0\textwidth]{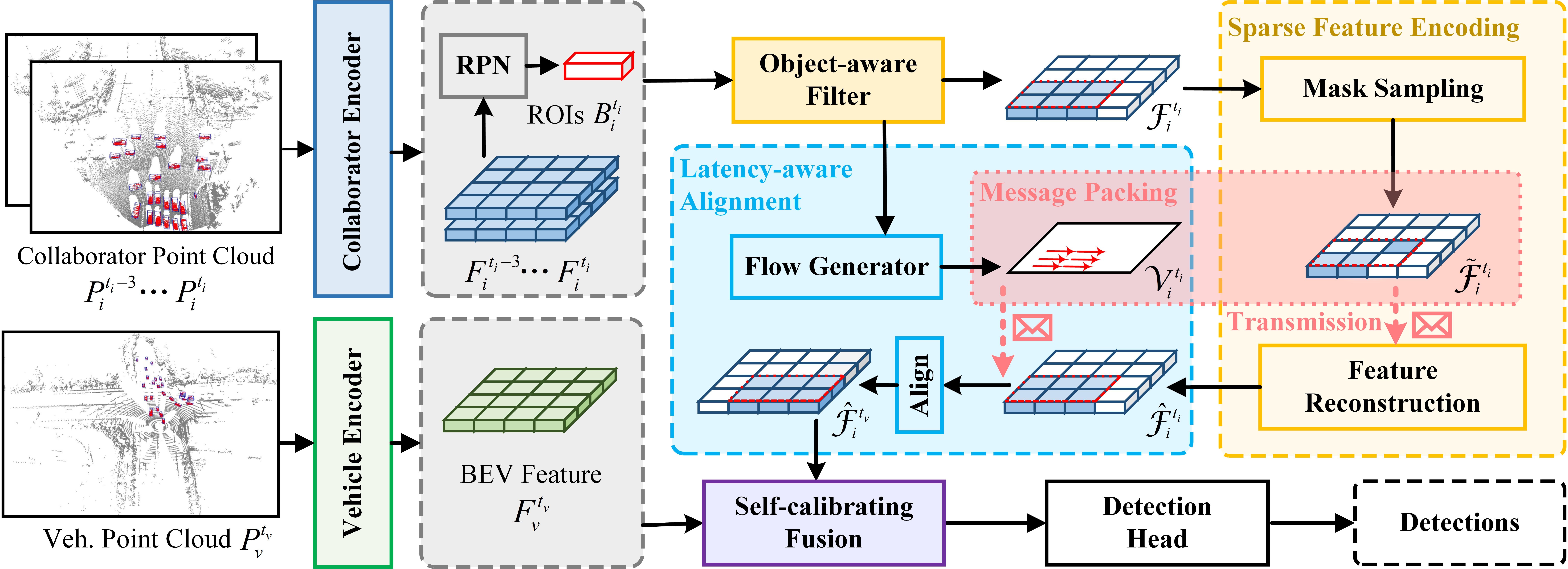}
\caption{\textbf{The overall framework of Sparse2comm.} The collaborator and ego encoders extract BEV features from LiDAR point clouds. \textbf{Sparse Feature Encoding} transmits masked sparse features and reconstructs complete semantics to minimize communication cost and resist packet loss. \textbf{Latency-Aware Alignment} predicts motion flow to temporally align delayed messages. \textbf{Self-Calibrating Fusion} self-supervisively estimates residual spatial offsets to refine cross-agent spatial alignment. The fused BEV features are finally decoded by a detection head for 3D object detection.}
 \label{fig:framework}
 \vspace{-0.2cm}
\end{figure*}

%% file: latex/fig/motivation_sparse_feature_encoding.tex
\begin{figure}[h!t]
\centering
\includegraphics[width=0.48\textwidth]{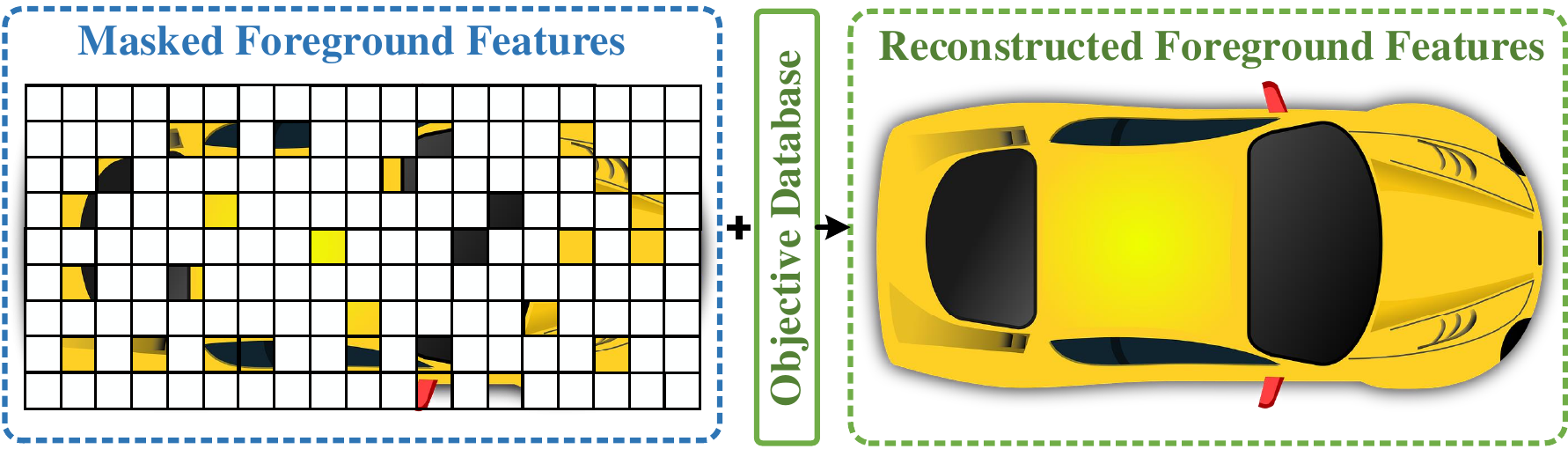}
\caption{\textbf{The Motivation of Sparse Feature Encoding.} Each object can be reconstructed from partial local features by leveraging a learned database of global priors, enabling complete semantic recovery under limited bandwidth.}
 \label{fig:motivation_sparse_feature_encoding}
\end{figure}

%% file: latex/fig/sparse_feature_encoding.tex
\begin{figure}[h!t]
	\centering
	\includegraphics[width=0.48\textwidth]{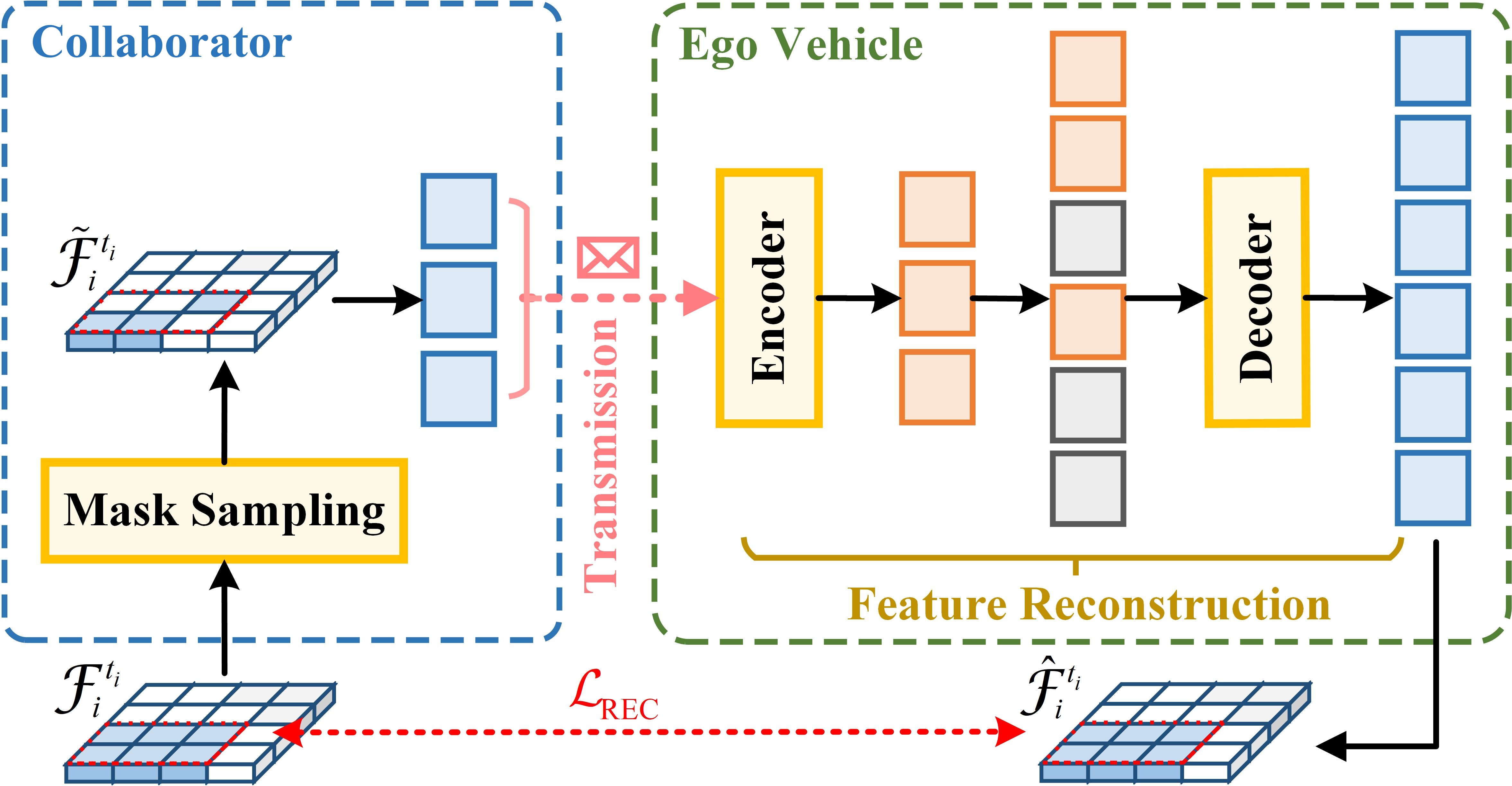}
	\caption{\textbf{Sparse Feature Encoding.} The collaborating agent performs Mask Sampling to transmit sparse object-foreground features under limited bandwidth, while the ego vehicle uses an encoder–decoder network for Feature Reconstruction to recover complete BEV features supervised by $\mathcal{L}_{\text{REC}}$.  
    %This design ensures efficient and robust communication against packet loss.
    }
 \label{fig:sparse_feature_encoding}
\end{figure}

%% file: latex/fig/latency_aware_alignment.tex
\begin{figure}[t]
	\centering
	\includegraphics[width=0.48\textwidth]{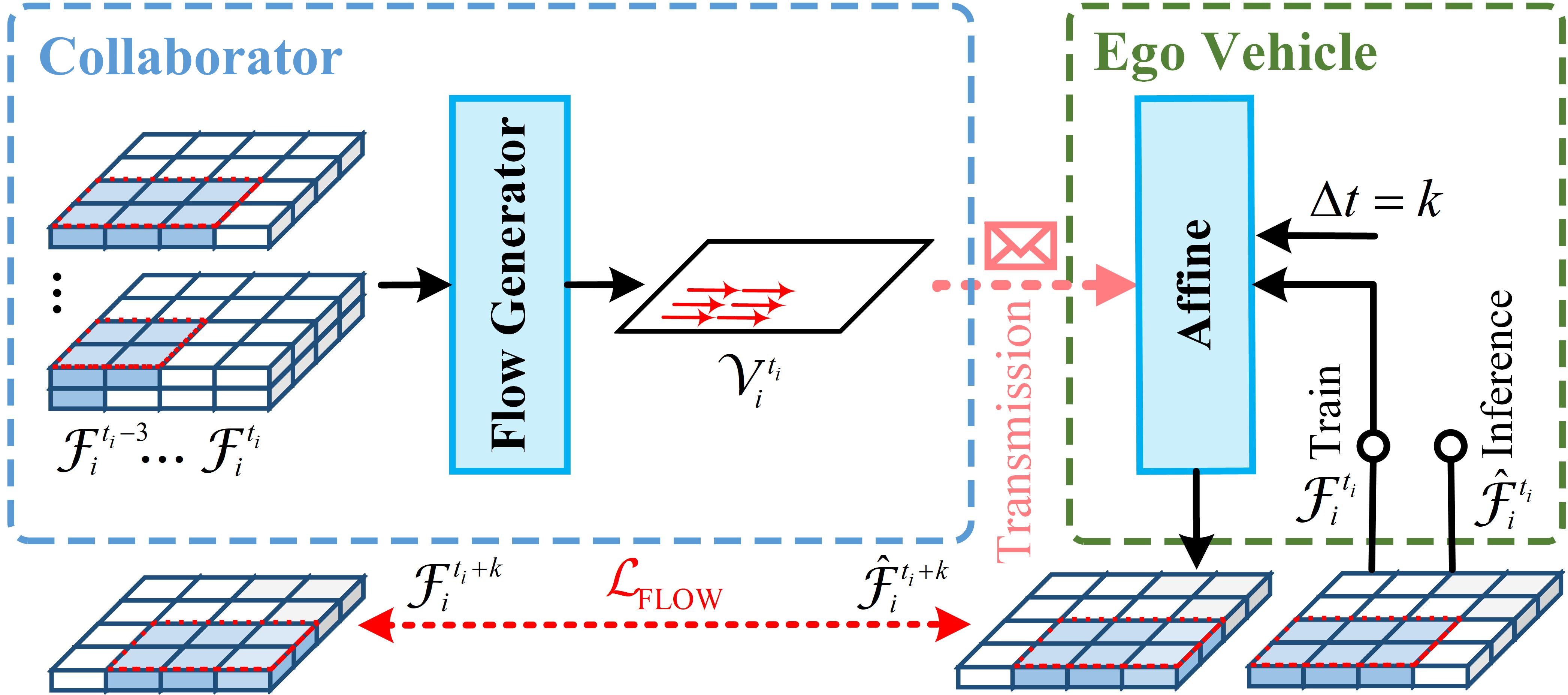}
	\caption{\textbf{Latency-aware Alignment.} Flow Generator predicts motion flow $\mathcal{V}_i^{t_i}$ from temporal BEV features on the collaborator side, which is transmitted to the ego vehicle for Affine-based temporal alignment. The module is trained with flow loss $\mathcal{L}_{\text{FLOW}}$ using future frames to ensure robustness against transmission delay.}
 \label{fig:latency_aware_alignment}
 \vspace{-0.1cm}
\end{figure}

%% file: latex/fig/self_calibrating_fusion.tex
\begin{figure}[t]
\centering
\includegraphics[width=0.48\textwidth]{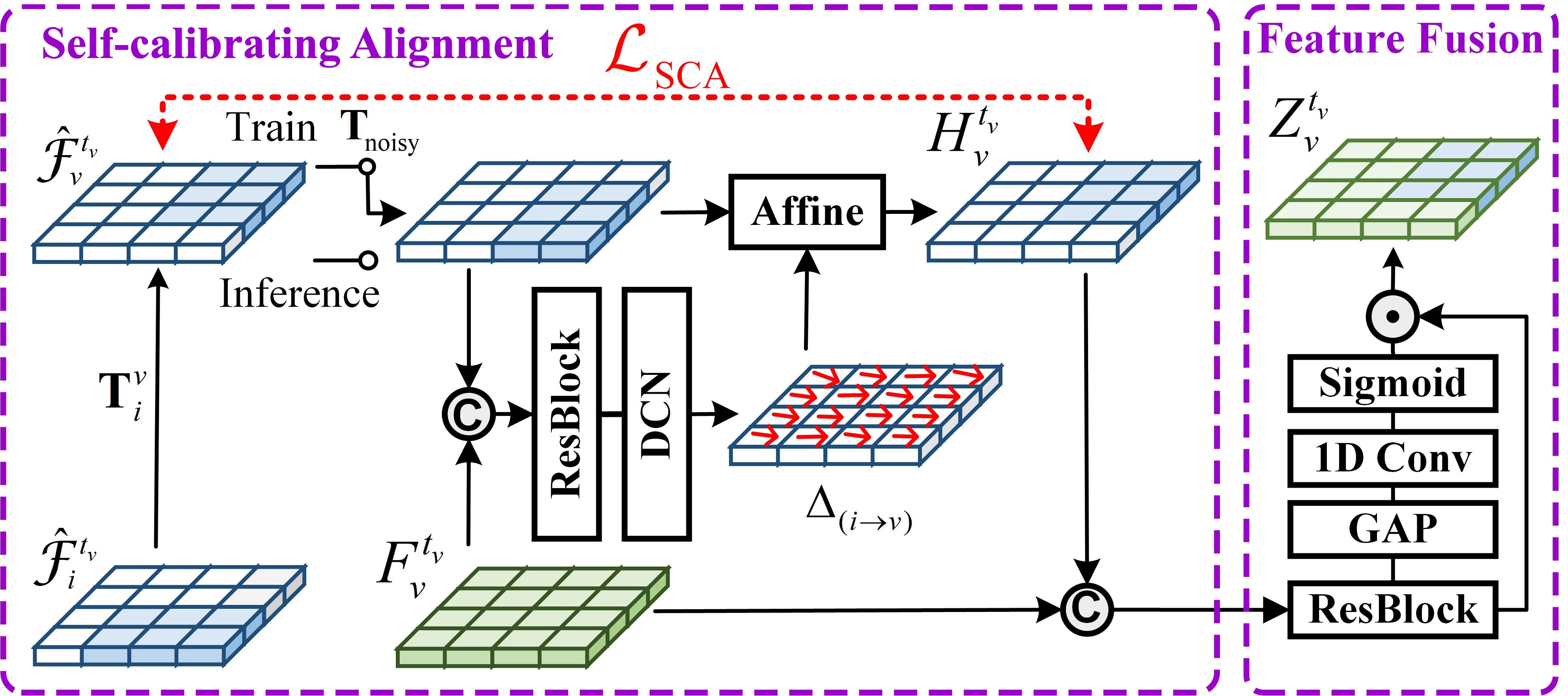}
\caption{\textbf{Self-calibrating Fusion.} Self-calibrating Alignment predicts pixel-wise offsets $\Delta_{(i\to v)}$ via residual and deformable convolutions to correct pose errors, guided by self-supervised loss $\mathcal{L}_{\text{SCA}}$. Feature Fusion further integrates aligned features to generate the final BEV representation $Z_v^{t_v}$.}
\label{fig:self_calibrating_fusion}
\vspace{-0.2cm}
\end{figure}

%% file: latex/sec/4_exps.tex
\section{Experiments}
\subsection{Datasets}

\input{latex/tab/exps_sota_multidataset}

We evaluate Sparse2comm on three cooperative perception datasets: DAIR-V2X~\cite{yu2022dair}, OpenV2V~\cite{xu2022opv2v}, and V2V4Real~\cite{xu2023v2v4real}. These datasets cover complementary cooperative settings, including vehicle--infrastructure cooperation, simulated vehicle--vehicle cooperation, and real-world vehicle--vehicle cooperation, allowing us to examine robustness beyond a single benchmark.

\noindent \textbf{DAIR-V2X}~\cite{yu2022dair} is a large-scale vehicle--infrastructure cooperative perception dataset collected from real urban scenarios. It contains synchronized vehicle-side and infrastructure-side LiDAR-camera data with 3D annotations, and introduces the VIC3D task where detection must account for both temporal asynchrony and communication cost. We use DAIR-V2X as the primary benchmark for evaluating vehicle--infrastructure cooperative 3D object detection under bandwidth and robustness constraints.

\noindent \textbf{OpenV2V}~\cite{xu2022opv2v} is an open simulated vehicle--vehicle cooperative perception benchmark built with CARLA and OpenCDA. It provides multi-agent sensor observations, accurate poses, and cooperative 3D annotations across diverse traffic scenarios, making it suitable for controlled evaluation of vehicle--vehicle feature fusion. We use OpenV2V to examine whether Sparse2comm generalizes to multi-agent V2V cooperation under a standardized simulation environment.

\noindent \textbf{V2V4Real}~\cite{xu2023v2v4real} is a large-scale real-world vehicle--vehicle cooperative perception dataset collected by two vehicles equipped with multi-modal sensors. It covers 410 km of driving routes and provides 20K LiDAR frames, 40K RGB images, 240K annotated 3D bounding boxes across five classes, and HD maps. We use its cooperative 3D object detection task to test whether Sparse2comm generalizes to real vehicle--vehicle cooperation with moving agents on both sides.

For all datasets, detection performance is evaluated using Average Precision (AP) at IoU thresholds of 0.5 and 0.7. To assess robustness under practical cooperative perception degradations, we use the same perturbation protocol across methods: (i) packet loss is simulated by randomly masking transmitted features with a ratio $r\sim\mathcal{U}(0,0.9)$; (ii) transmission delay is simulated by using delayed cooperative features, corresponding to a latency of $200$ ms; and (iii) spatial misalignment is simulated by adding rotational noise $\Delta R\sim\mathcal{N}(0,3.0^\circ)$ and translational noise $\Delta T \sim\mathcal{N}(0,0.5)$ m. All compared methods are evaluated under the same feature-transmission protocol without additional compression for fair comparison.

\subsection{Implementation Details}
Sparse2comm uses the same sparse feature communication interface, progressive restoration modules, and robustness perturbation protocol across all datasets, with dataset-specific detection ranges and evaluation protocols following the corresponding benchmarks. On DAIR-V2X, we set the point cloud range to $[-100.8,100.8]\times[-40,40]\times[-5,3]$ meters and the voxel size to $0.4\times0.4\times4$ meters, and evaluate only the Car class. On OpenV2V, we follow the standard LiDAR-based vehicle--vehicle cooperative detection setting, use PointPillars~\cite{lang2019pointpillars} as the BEV encoder, and evaluate the vehicle category under the official protocol. On V2V4Real, we use the LiDAR-based PointPillars backbone, set the detection region to $[-100,100]$ meters along the $x$ axis and $[-40,40]$ meters along the $y$ axis, merge all vehicle types into one vehicle class, and report AP at IoU thresholds of $0.5$ and $0.7$. The Sync and Async settings are further included in our unified robustness protocol for packet loss, communication delay, and spatial misalignment. The training follows three stages: collaborator encoder and RPN pre-training, Sparse Feature Encoding pre-training, and end-to-end cooperative fine-tuning. We use Adam~\cite{kingma2014adam} with a learning rate of $0.01$, a batch size of $6$, and $80$ epochs. For V2V4Real, we train for $60$ epochs using AdamW with an initial learning rate of $0.001$, a batch size of $4$ per GPU, a weight decay of $1\times10^{-2}$, cosine learning-rate decay, and standard point-cloud augmentations.

\input{latex/tab/exps_robustness_dair}
\input{latex/fig/robustness_different_conditions}

\subsection{Quantitative Results}
\subsubsection{Comparison with State-of-the-art} 
We benchmark Sparse2comm on DAIR-V2X~\cite{yu2022dair}, OpenV2V~\cite{xu2022opv2v}, and V2V4Real~\cite{xu2023v2v4real}, covering dense feature-level cooperation~\cite{chen2019f,mehr2019disconet,wang2020v2vnet,xu2022v2x}, communication-efficient methods~\cite{liu2020when2com,hu2022where2comm,yang2023how2comm,xu2025cosdh}, and robustness-oriented methods~\cite{wei2023asynchrony,10160546,xu2025codyntrust}. Under clean communication, Sparse2comm obtains 69.42/53.78 AP@0.5/AP@0.7 on DAIR-V2X, 93.70/87.66 on OpenV2V, and 71.46/45.36 on V2V4Real. It ranks first on DAIR-V2X, second on OpenV2V behind CoSDH~\cite{xu2025cosdh} by only 0.50/0.53 AP, and remains competitive on V2V4Real, showing that sparse communication does not noticeably compromise normal detection accuracy while using only 1.0\% relative bandwidth.

The advantage becomes much clearer under the mixed setting, where packet loss, transmission delay, and spatial misalignment occur together. Sparse2comm reaches 58.10/39.54, 88.28/74.53, and 56.18/33.15 AP@0.5/AP@0.7 on the three datasets, improving over the strongest prior results by +15.75/+9.37, +7.13/+4.11, and +11.11/+7.34, respectively. Many dense feature-level baselines drop close to or below No Fusion once cooperative messages are corrupted, while communication-efficient and robustness-oriented methods remain less balanced under coupled disturbances. In contrast, Sparse2comm preserves SOTA-level clean accuracy and achieves the best mixed-setting performance across all datasets, providing a stronger trade-off among accuracy, robustness, and communication efficiency.

\subsubsection{Robustness to Real-world Challenges}
We then isolate the three major robustness factors on DAIR-V2X~\cite{yu2022dair}: packet loss, transmission delay, and spatial misalignment. Table~\ref{tab:robustness_dair} gives the numerical breakdown, while Fig.~\ref{fig:robustness_different_conditions} traces performance as each disturbance becomes stronger. Sparse2comm achieves the strongest overall robustness across the evaluated disturbances. Under packet loss, it obtains 67.12/50.77 AP@0.5/AP@0.7, outperforming the second-best method, CoDynTrust~\cite{xu2025codyntrust}, by +10.20/+7.19. Under spatial misalignment, Sparse2comm reaches 63.08/45.39, improving over CoAlign~\cite{10160546} by +4.06/+1.72, even though CoAlign is specifically designed for pose alignment. Under transmission delay, Sparse2comm achieves the best AP@0.5 and remains close to the best AP@0.7 result, indicating that it maintains competitive temporal robustness while being more balanced across other disturbances.

The key observation is that existing methods tend to be robust to only one degradation type. How2comm~\cite{yang2023how2comm}, CoBEVFlow~\cite{wei2023asynchrony}, and CoDynTrust~\cite{xu2025codyntrust} are strong under temporal asynchrony, and CoAlign~\cite{10160546} is effective under spatial misalignment, but their advantages do not consistently transfer to packet loss or coupled disturbances. The degradation curves make this imbalance clearer: many dense feature-level~\cite{chen2019f,mehr2019disconet,wang2020v2vnet,xu2022v2x} and communication-efficient methods~\cite{liu2020when2com,hu2022where2comm,yang2023how2comm,xu2025cosdh} degrade rapidly as disturbance intensity increases, and some even approach or fall below the No Fusion baseline. In contrast, Sparse2comm exhibits a slower performance decay across all three curves and achieves 58.10/39.54 under the mixed setting, surpassing the strongest prior result, CoDynTrust~\cite{xu2025codyntrust}, by +15.75/+9.37 with only 2.6 Mbps bandwidth. Taken together, the targeted and mixed results support Sparse2comm as a more balanced robustness solution for practical cooperative perception.

\subsection{Ablation Studies}
\subsubsection{Effectiveness of Major Components} The component ablation in Tab.~\ref{tab:module_ablation} isolates the role of SFE, LAA, and SCF in the full framework. When all three modules are combined, Sparse2comm achieves the best performance, yielding +20.15 AP@0.5 over the baseline under noisy conditions. This confirms the complementary roles of the three modules in enhancing both accuracy and robustness across diverse real-world challenges.
\subsubsection{Impact of Training Strategy}
The training ablation in Tab.~\ref{tab:training_ablation} favors the proposed multi-stage strategy over direct joint training, which optimizes all components simultaneously. Gradually pre-training key modules before end-to-end fine-tuning leads to more stable optimization and stronger generalization to real-world perturbations.
\subsubsection{Impact of Sampling Strategy}
For feature sampling, Object-aware Sampling outperforms both Uniform and Random Sampling in Tab.~\ref{tab:masking_ablation}. Here, Uniform Sampling denotes evenly spaced sampling, while Random Sampling assigns equal probability to all regions. By focusing on object-centric areas during feature masking, our strategy preserves more discriminative information and effectively improves both detection accuracy and robustness.

\input{latex/tab/exps_ablation_modules}
\input{latex/tab/exps_ablation_multi_stage_training}
\input{latex/tab/exps_ablation_sampling}
\input{latex/tab/exp_ablation_data_augmentation}
\input{latex/tab/exp_sota_efficiency}
\input{latex/tab/exp_ablation_efficiency}
\input{latex/tab/exp_ablation_sfe_config}

\subsubsection{Beyond Data Augmentation}
To verify that the performance gains of Sparse2comm do not originate from data augmentation,
we evaluate the effects of random masking (RM) and pose disturbance (PD) on both the baseline and our model.
In the data-augmentation control of Table~\ref{tab:data_aug_ablation}, introducing RM or PD consistently degrades the baseline under the ideal setting. RM slightly reduces accuracy
(from 67.68 to 66.41 AP@0.5 and from 52.18 to 49.92 AP@0.7), while PD causes a larger drop 
(64.56 AP@0.5 and 47.58 AP@0.7). Combining RM+PD further lowers performance to 
67.03 AP@0.5 and 49.86 AP@0.7. Under the noisy setting, RM and PD bring only marginal improvements, and RM+PD yields 
45.47 AP@0.5 and 28.43 AP@0.7. However, these gains remain insufficient to mitigate the 
performance degradation observed on the baseline. In contrast, Sparse2comm achieves \textbf{69.42 / 53.78} AP@0.5/0.7 under the ideal setting,
outperforming the non-augmented baseline by \textbf{+1.74 / +1.60}. Under noise, Sparse2comm reaches
\textbf{58.10 / 39.54}, surpassing the corresponding baseline by \textbf{+20.15 / +11.79}.
This comparison supports that the improvements of Sparse2comm arise from its architectural design 
rather than augmentation heuristics, effectively ruling out data augmentation as a confounding factor.

\subsubsection{Efficiency Analysis}
For deployment efficiency, we profile Sparse2comm and the representative baselines under a shared DAIR-V2X setting, using the same machine, BEV range, and inference-stage configuration as in Table~\ref{tab:efficiency_comparison}. In addition to model-side parameters, FLOPs, and pure inference latency, we explicitly account for the communication time required to transmit intermediate cooperative features. Given the transmitted bandwidth $\mathrm{BW}$ in Mbps, evaluation frequency $f{=}10$ Hz, and V2X link rate $R$, the transmission time is computed as $T_{\mathrm{tx}}(R)=1000\mathrm{BW}/(fR)$, and the end-to-end latency is $T_{\mathrm{e2e}}=T_{\mathrm{model}}+T_{\mathrm{tx}}$. We report both a conservative 10 Mbps link and an optimistic 50 Mbps link to reflect practical and high-capacity deployment conditions.

Model-only latency can substantially underestimate the deployment cost of dense feature-level cooperation. Although several full-feature or alignment-based methods have moderate inference latency, their dense communication cost leads to second-level end-to-end latency under a 10 Mbps V2X link. Communication-efficient methods reduce this cost, but their transmitted feature volume remains noticeably higher than Sparse2comm. In contrast, Sparse2comm requires only 2.6 Mbps on DAIR-V2X, reducing the transmission time to 26.0 ms under the conservative 10 Mbps link and 5.2 ms under the 50 Mbps link. Consequently, Sparse2comm achieves 95.7 ms end-to-end latency under 10 Mbps and 74.9 ms under 50 Mbps, while maintaining strong robustness in Table~\ref{tab:robustness_dair}. The component-level profile in Table~\ref{tab:efficiency_ablation} further breaks down the computational cost: starting from the object-foreground baseline, SFE, LAA, and SCF each introduce controlled overhead, and the complete system increases latency from 51.2 ms to 69.7 ms. Overall, Sparse2comm offers a favorable system-level trade-off between computation, communication, robustness, and real-time deployment feasibility.

\subsubsection{SFE Architecture}
We further vary the SFE encoder--decoder depth in Tab.~\ref{tab:exp_ablation_sfe_config}, where $N_E$ and $N_D$ denote the number of ViT~\cite{dosovitskiy2020vit} blocks in the encoder and decoder, respectively. This ablation evaluates shallow, encoder-heavy, decoder-heavy, and deeper configurations to understand how capacity allocation influences performance under both ideal and noisy conditions.

The alternative SFE designs, including the shallow (S1), encoder-heavy (S2), decoder-heavy (S4), and deeper configuration (S5), do not provide the same balance across clean and noisy settings. Shallow and unbalanced variants either lack sufficient representational capacity or fail to utilize it efficiently, and simply increasing depth does not resolve the robustness issue, as S5 still performs poorly under noise. In contrast, the balanced architecture (S3 with $N_E{=}5$ and $N_D{=}2$) obtains the highest accuracy in both ideal and noisy scenarios, reaching \textbf{69.42 / 53.78} under ideal conditions and \textbf{58.10 / 39.54} under noisy conditions. This pattern suggests that allocating moderate depth to both the encoder and the decoder provides the most effective and robust design for Sparse Feature Encoding.

\input{latex/fig/visualization}
\input{latex/fig/sfe_module_visualization}

\subsection{Qualitative Results}

\subsubsection{Detection Result Visualization}
Figs.~\ref{fig:qualitative_results} and~\ref{fig:qualitative_v2v4real} show representative noisy-condition detections on DAIR-V2X and V2V4Real, respectively. Green/red boxes denote ground-truth/predicted objects, and magenta dashed regions highlight challenging areas.
On DAIR-V2X~\cite{yu2022dair}, baseline methods often miss compact vehicle clusters or produce spatially shifted boxes, especially for adjacent and partially occluded targets. Sparse2comm detects more vehicles in these crowded regions and aligns the predicted boxes more closely with the ground truth.
On V2V4Real~\cite{xu2023v2v4real}, noisy communication causes unstable long-range detections and inaccurate orientations along the driving corridor. In contrast, Sparse2comm produces fewer missed detections and more reliable box placement, showing its effectiveness in real-world vehicle--vehicle cooperation.
These comparisons are consistent with Tables~\ref{tab:sota_multidataset} and~\ref{tab:robustness_dair}, where Sparse2comm improves noisy-condition detection by restoring semantic content, temporal consistency, and spatial alignment before fusion.

\subsubsection{SFE Reconstruction Visualization}
To inspect the reconstruction process, Fig.~\ref{fig:sfe_visualization} visualizes refined, mask-sampled, and reconstructed BEV features produced by Sparse Feature Encoding 
(SFE). For each packet--loss ratio (10\%--100\%), we include: (i) the refined BEV features 
$\mathcal{F}_i^{t_i}$, (ii) the mask-sampled BEV features $\tilde{\mathcal{F}}_i^{t_i}$ produced by the Mask 
Sampling module, (iii) the reconstructed BEV features $\hat{\mathcal{F}}_i^{t_i}$ generated by the Feature 
Reconstruction module, and (iv-vi) the corresponding information-loss maps and loss histograms.

The mask-sampled BEV features simulate the actual information received by a vehicle under packet loss. 
As the packet-loss ratio increases, $\tilde{\mathcal{F}}_i^{t_i}$ exhibits severe degradation, and the 
mask-sampled loss $|\mathcal{F}_i^{t_i} - \tilde{\mathcal{F}}_i^{t_i}|^2$ clearly highlights large missing 
regions. In contrast, the reconstruction module effectively recovers the missing content, producing features structurally close to the refined BEV representation. The reconstructed loss $|\mathcal{F}_i^{t_i} - \hat{\mathcal{F}}_i^{t_i}|^2$ remains near zero, and the corresponding histogram 
distributions collapse sharply toward zero across all packet-loss settings.
This behavior is consistent even under extremely high packet loss (e.g., 90\%), demonstrating 
the strong capability of SFE to restore semantic structure from severely degraded inputs. The near-zero reconstruction loss supports that SFE not only compensates for transmission-induced information loss in real cooperative communication, but also enables an efficient sparse-to-dense reconstruction paradigm. 
In practice, collaborating agents can transmit sparsely sampled features with substantially reduced bandwidth 
requirements, while the ego vehicle reconstructs dense features with minimal performance degradation.

%% file: latex/tab/exps_sota_multidataset.tex
\begin{table*}[t]
\setlength{\abovecaptionskip}{8pt}
\renewcommand{\arraystretch}{1.15}
\setlength{\tabcolsep}{2.2pt}
\caption{\textbf{Main comparison with representative cooperative perception methods across datasets.}
Each performance cell reports AP@0.5 / AP@0.7. ``Clean'' denotes evaluation without artificial communication or pose perturbation, while ``Mixed'' denotes the unified robustness setting with packet loss, transmission delay, and spatial misalignment. BW is reported as the relative transmitted feature volume with respect to dense feature-level cooperation on each dataset, where dense feature transmission is normalized to 100\%. Gray entries denote the single-agent No Fusion baseline, while black bold and underlined numbers indicate the best and second-best cooperative-method results in each column, respectively.}
\label{tab:sota_multidataset}
\scriptsize
\centering
\begin{tabularx}{1.0\textwidth}{l l *{6}{>{\centering\arraybackslash}X} c}
\toprule
\multirow{2}{*}{\textbf{Method}}
& \multirow{2}{*}{\textbf{Category}}
& \multicolumn{2}{c}{\textbf{DAIR-V2X}}
& \multicolumn{2}{c}{\textbf{OpenV2V}}
& \multicolumn{2}{c}{\textbf{V2V4Real}}
& \multirow{2}{*}{\textbf{BW (\%)$\downarrow$}} \\
\cmidrule(lr){3-4}\cmidrule(lr){5-6}\cmidrule(lr){7-8}
& & \textbf{Clean} & \textbf{Mixed} & \textbf{Clean} & \textbf{Mixed} & \textbf{Clean} & \textbf{Mixed} & \\
\midrule
\multicolumn{9}{l}{\textit{Basic and classic cooperative perception}} \\
\textcolor{gray}{No Fusion} & \textcolor{gray}{Single-agent} & \textcolor{gray}{42.45 / 30.78} & \textcolor{gray}{42.45 / 30.78} & \textcolor{gray}{79.78 / 67.16} & \textcolor{gray}{79.78 / 67.16} & \textcolor{gray}{40.31 / 22.47} & \textcolor{gray}{40.31 / 22.47} & \textcolor{gray}{0} \\
F-Cooper~\cite{chen2019f} & Full feature & 49.83 / 34.21 & 29.32 / 22.74 & 88.15 / 80.97 & 68.54 / 55.62 & 66.08 / 35.21 & 37.41 / 17.42 & 100.0 \\
DiscoNet~\cite{mehr2019disconet} & Full feature & 51.31 / 35.77 & 33.26 / 27.50 & 88.90 / 82.03 & 70.26 / 58.34 & 65.47 / 35.96 & 39.60 / 19.05 & 100.0 \\
V2VNet~\cite{wang2020v2vnet} & Full feature & 61.71 / 46.25 & 33.95 / 21.71 & 92.16 / 85.38 & 69.58 / 56.71 & 67.14 / 36.42 & 38.38 / 17.79 & 100.0 \\
V2X-ViT~\cite{xu2022v2x} & Full feature & 58.37 / 42.77 & 37.31 / 26.41 & 91.43 / 84.28 & 73.48 / 61.26 & 68.36 / 39.08 & 41.77 / 20.99 & 100.0 \\
\midrule
\multicolumn{9}{l}{\textit{Robust cooperative perception}} \\
CoBEVFlow~\cite{wei2023asynchrony} & Temporal alignment & 66.91 / 51.64 & 37.16 / 27.20 & 92.23 / 85.83 & 80.36 / 69.04 & 69.08 / 43.72 & 41.81 / 21.16 & 5.6 \\
CoAlign~\cite{10160546} & Pose alignment & 68.72 / 53.06 & 40.76 / 29.68 & 92.94 / 86.56 & 76.93 / 65.42 & \textbf{72.71} / 45.04 & 44.47 / 24.84 & 100.0 \\
CoDynTrust~\cite{xu2025codyntrust} & Asynchronous robustness & 67.97 / 52.36 & \underline{42.35 / 30.17} & 93.28 / 87.05 & \underline{81.15 / 70.42} & 71.67 / 44.93 & \underline{45.07 / 25.81} & 10.6 \\
\midrule
\multicolumn{9}{l}{\textit{Communication-efficient cooperative perception}} \\
When2comm~\cite{liu2020when2com} & Adaptive comm. & 54.96 / 36.15 & 36.89 / 26.33 & 89.47 / 82.15 & 72.35 / 60.54 & 64.23 / 34.71 & 40.49 / 19.59 & 100.0 \\
How2comm~\cite{yang2023how2comm} & Pragmatic comm. & 68.76 / 53.08 & 41.44 / 29.43 & 93.47 / 87.44 & 78.25 / 66.87 & 71.63 / \underline{45.74} & 44.07 / 23.88 & 3.4 \\
CoSDH~\cite{xu2025cosdh} & Hybrid fusion & \underline{68.91 / 53.14} & 40.62 / 29.01 & \textbf{94.20 / 88.19} & 79.64 / 68.15 & \underline{72.42} / \textbf{46.09} & 43.01 / 22.98 & \underline{2.8} \\
Where2comm~\cite{hu2022where2comm} & Selective feature & 67.68 / 52.18 & 37.95 / 27.75 & 93.00 / 86.97 & 75.62 / 63.46 & 70.24 / 44.63 & 40.82 / 20.54 & 5.0 \\
\midrule
\textbf{Sparse2comm (Ours)} & Progressive restoration & \textbf{69.42 / 53.78} & \textbf{58.10 / 39.54} & \underline{93.70 / 87.66} & \textbf{88.28 / 74.53} & 71.46 / 45.36 & \textbf{56.18 / 33.15} & \textbf{1.0} \\
\bottomrule
\end{tabularx}
\vspace{-0.1cm}
\end{table*}

%% file: latex/tab/exps_robustness_dair.tex
\begin{table*}[h!t]
\setlength{\abovecaptionskip}{8pt}
\renewcommand{\arraystretch}{1.15}
\setlength{\tabcolsep}{1.2pt}
\caption{\textbf{Targeted robustness comparison on DAIR-V2X.}
All entries report AP@0.5 / AP@0.7, and the method list follows the main comparison in Table~\ref{tab:sota_multidataset}. Packet Loss, Transmission Delay, and Spatial Misalignment denote representative robustness settings evaluated under the same protocol for all methods. Mixed combines packet loss, transmission delay, and spatial misalignment to evaluate robustness under coupled real-world degradations. Since all results are evaluated on DAIR-V2X with the same feature shape, BW is reported as the transmitted bandwidth in Mbps. Gray entries denote the single-agent No Fusion baseline, while black bold and underlined numbers indicate the best and second-best cooperative-method results in each column, respectively.}
\label{tab:robustness_dair}
\scriptsize
\centering
\begin{tabularx}{1.0\textwidth}{@{}l l *{5}{>{\centering\arraybackslash}X} c@{}}
\toprule
\textbf{Method}
& \textbf{Category}
& \textbf{Clean}
& \textbf{Packet Loss}
& \textbf{Delay}
& \textbf{Misalignment}
& \textbf{Mixed}
& \textbf{BW$\downarrow$ (Mbps)} \\
\midrule
\multicolumn{8}{l}{\textit{Basic and classic cooperative perception}} \\
\textcolor{gray}{No Fusion} & \textcolor{gray}{Single-agent} & \textcolor{gray}{42.45 / 30.78} & \textcolor{gray}{42.45 / 30.78} & \textcolor{gray}{42.45 / 30.78} & \textcolor{gray}{42.45 / 30.78} & \textcolor{gray}{42.45 / 30.78} & \textcolor{gray}{0} \\
F-Cooper~\cite{chen2019f} & Full feature & 49.83 / 34.21 & 43.17 / 29.63 & 42.58 / 29.37 & 38.52 / 27.13 & 29.32 / 22.74 & 258.1 \\
DiscoNet~\cite{mehr2019disconet} & Full feature & 51.31 / 35.77 & 46.30 / 32.27 & 44.72 / 31.55 & 41.26 / 30.94 & 33.26 / 27.50 & 258.1 \\
V2VNet~\cite{wang2020v2vnet} & Full feature & 61.71 / 46.25 & 48.83 / 34.49 & 57.72 / 34.83 & 43.63 / 27.28 & 33.95 / 21.71 & 258.1 \\
V2X-ViT~\cite{xu2022v2x} & Full feature & 58.37 / 42.77 & 48.02 / 35.68 & 54.44 / 34.04 & 45.59 / 30.36 & 37.31 / 26.41 & 258.1 \\
\midrule
\multicolumn{8}{l}{\textit{Robust cooperative perception}} \\
CoBEVFlow~\cite{wei2023asynchrony} & Temporal alignment & 66.91 / 51.64 & 52.64 / 41.36 & 60.28 / 42.36 & 48.32 / 31.18 & 37.16 / 27.20 & 14.5 \\
CoAlign~\cite{10160546} & Pose alignment & 68.72 / 53.06 & 54.72 / 42.06 & 54.06 / 39.84 & \underline{59.02 / 43.67} & 40.76 / 29.68 & 258.1 \\
CoDynTrust~\cite{xu2025codyntrust} & Asynchronous robustness & 67.97 / 52.36 & \underline{56.92 / 43.58} & \underline{64.58} / 47.56 & 53.57 / 38.56 & \underline{42.35 / 30.17} & 27.4 \\
\midrule
\multicolumn{8}{l}{\textit{Communication-efficient cooperative perception}} \\
When2comm~\cite{liu2020when2com} & Adaptive comm. & 54.96 / 36.15 & 46.32 / 31.40 & 52.48 / 33.11 & 45.02 / 29.30 & 36.89 / 26.33 & 258.1 \\
How2comm~\cite{yang2023how2comm} & Pragmatic comm. & 68.76 / 53.08 & 55.86 / 42.94 & 64.41 / \textbf{48.13} & 56.24 / 42.16 & 41.44 / 29.43 & 8.8 \\
CoSDH~\cite{xu2025cosdh} & Hybrid fusion & \underline{68.91 / 53.14} & 56.34 / 43.28 & 56.84 / 42.07 & 50.96 / 35.94 & 40.62 / 29.01 & \underline{7.2} \\
Where2comm~\cite{hu2022where2comm} & Selective feature & 67.68 / 52.18 & 53.00 / 41.80 & 58.97 / 41.42 & 48.90 / 31.51 & 37.95 / 27.75 & 12.9 \\
\midrule
\textbf{Sparse2comm (Ours)} & Progressive restoration & \textbf{69.42 / 53.78} & \textbf{67.12 / 50.77} & \textbf{64.73} / \underline{47.94} & \textbf{63.08 / 45.39} & \textbf{58.10 / 39.54} & \textbf{2.6} \\
\bottomrule
\end{tabularx}
\end{table*}

%% file: latex/fig/robustness_different_conditions.tex
\begin{figure*}[t]
\centering
\setlength{\abovecaptionskip}{3pt}
\definecolor{methodgray}{RGB}{115,115,115}
\definecolor{methodpurple}{RGB}{145,96,190}
\definecolor{methodorange}{RGB}{232,139,45}
\definecolor{methodgreen}{RGB}{54,154,111}
\definecolor{methodteal}{RGB}{65,127,145}
\definecolor{methodblue}{RGB}{62,139,217}
\definecolor{methodmagenta}{RGB}{188,91,150}
\definecolor{methodolive}{RGB}{133,154,45}
\definecolor{methodcyan}{RGB}{52,164,188}
\definecolor{methodbrown}{RGB}{149,104,72}
\definecolor{methodviolet}{RGB}{110,108,196}
\definecolor{methodred}{RGB}{211,48,48}
\pgfplotsset{
robust axis/.style={
width=0.34\textwidth,
height=0.205\textwidth,
grid=major,
grid style={dashed,gray!35},
xminorticks=false,
yminorticks=false,
tick label style={font=\scriptsize},
label style={font=\scriptsize},
title style={font=\scriptsize,yshift=-0.7ex},
legend style={
font=\scriptsize,
draw=black,
fill=white,
legend columns=7,
/tikz/every even column/.append style={column sep=0.18cm}
},
line width=0.7pt,
mark size=1.5pt,
},
nf/.style={black,mark=*,mark options={solid}},
fc/.style={methodgray,mark=square*,mark options={solid}},
dn/.style={methodpurple,mark=triangle*,mark options={solid}},
v2vn/.style={methodorange,mark=diamond*,mark options={solid}},
v2xvit/.style={methodgreen,mark=pentagon*,mark options={solid}},
when/.style={methodteal,mark=triangle*,mark options={solid,rotate=180}},
where/.style={methodblue,mark=*,mark options={solid}},
how/.style={methodmagenta,mark=diamond*,mark options={solid}},
cosdh/.style={methodolive,mark=square*,mark options={solid}},
cobev/.style={methodcyan,mark=otimes*,mark options={solid}},
coalign/.style={methodbrown,mark=star,mark options={solid}},
codyn/.style={methodviolet,mark=halfsquare*,mark options={solid}},
ours/.style={methodred,mark=square*,mark options={solid},line width=0.95pt},
}
\resizebox{\textwidth}{!}{%
\begin{tikzpicture}[font=\scriptsize]
\begin{groupplot}[
group style={group size=3 by 2,horizontal sep=0.042\textwidth,vertical sep=0.052\textwidth},
robust axis,
]
\nextgroupplot[
xlabel={Packet Loss Ratio (\%)},
ylabel={AP@0.5},
xmin=0,xmax=90,ymin=30,ymax=72,
xtick={0,10,30,50,70,90},
legend to name=robustlegend,
]
\addplot+[nf] coordinates {(0,42.45) (10,42.45) (30,42.45) (50,42.45) (70,42.45) (90,42.45)};
\addlegendentry{No Fusion}
\addplot+[fc] coordinates {(0,49.83) (10,48.10) (30,45.62) (50,43.17) (70,40.23) (90,36.74)};
\addlegendentry{F-Cooper}
\addplot+[dn] coordinates {(0,51.31) (10,49.64) (30,47.82) (50,46.30) (70,43.88) (90,40.96)};
\addlegendentry{DiscoNet}
\addplot+[v2vn] coordinates {(0,61.71) (10,59.03) (30,54.61) (50,48.83) (70,42.18) (90,35.62)};
\addlegendentry{V2VNet}
\addplot+[v2xvit] coordinates {(0,58.37) (10,56.42) (30,52.28) (50,48.02) (70,45.43) (90,41.76)};
\addlegendentry{V2X-ViT}
\addplot+[when] coordinates {(0,54.96) (10,53.08) (30,49.76) (50,46.32) (70,42.84) (90,38.61)};
\addlegendentry{When2comm}
\addplot+[where] coordinates {(0,67.68) (10,65.92) (30,60.84) (50,53.00) (70,47.26) (90,42.18)};
\addlegendentry{Where2comm}
\addplot+[how] coordinates {(0,68.76) (10,66.92) (30,62.38) (50,55.86) (70,50.42) (90,45.18)};
\addlegendentry{How2comm}
\addplot+[cosdh] coordinates {(0,68.91) (10,67.12) (30,62.84) (50,56.34) (70,51.06) (90,46.73)};
\addlegendentry{CoSDH}
\addplot+[cobev] coordinates {(0,66.91) (10,65.54) (30,60.26) (50,52.64) (70,47.05) (90,41.36)};
\addlegendentry{CoBEVFlow}
\addplot+[coalign] coordinates {(0,68.72) (10,66.18) (30,61.24) (50,54.72) (70,49.18) (90,44.63)};
\addlegendentry{CoAlign}
\addplot+[codyn] coordinates {(0,67.97) (10,66.28) (30,62.72) (50,56.92) (70,52.38) (90,48.26)};
\addlegendentry{CoDynTrust}
\addplot+[ours] coordinates {(0,69.42) (10,68.92) (30,68.02) (50,67.12) (70,65.16) (90,62.74)};
\addlegendentry{Sparse2comm}

\nextgroupplot[
xlabel={Transmission Delay (ms)},
xmin=0,xmax=500,ymin=30,ymax=72,
xtick={0,100,200,300,400,500},
]
\addplot+[nf] coordinates {(0,42.45) (100,42.45) (200,42.45) (300,42.45) (400,42.45) (500,42.45)};
\addplot+[fc] coordinates {(0,49.83) (100,42.58) (200,39.82) (300,37.68) (400,36.04) (500,34.71)};
\addplot+[dn] coordinates {(0,51.31) (100,44.72) (200,41.36) (300,39.18) (400,37.72) (500,36.41)};
\addplot+[v2vn] coordinates {(0,61.71) (100,57.72) (200,47.36) (300,42.58) (400,40.12) (500,38.29)};
\addplot+[v2xvit] coordinates {(0,58.37) (100,54.44) (200,47.82) (300,42.63) (400,40.80) (500,39.41)};
\addplot+[when] coordinates {(0,54.96) (100,52.48) (200,47.15) (300,42.46) (400,40.63) (500,39.18)};
\addplot+[where] coordinates {(0,67.68) (100,58.97) (200,53.24) (300,48.06) (400,45.62) (500,44.28)};
\addplot+[how] coordinates {(0,68.76) (100,64.41) (200,61.74) (300,59.12) (400,56.87) (500,54.96)};
\addplot+[cosdh] coordinates {(0,68.91) (100,56.84) (200,51.08) (300,47.22) (400,44.63) (500,42.74)};
\addplot+[cobev] coordinates {(0,66.91) (100,60.28) (200,54.16) (300,49.72) (400,46.81) (500,45.06)};
\addplot+[coalign] coordinates {(0,68.72) (100,54.06) (200,49.18) (300,46.43) (400,44.28) (500,42.85)};
\addplot+[codyn] coordinates {(0,67.97) (100,64.58) (200,61.08) (300,58.76) (400,56.98) (500,55.62)};
\addplot+[ours] coordinates {(0,69.42) (100,64.73) (200,61.68) (300,59.34) (400,57.82) (500,56.63)};

\nextgroupplot[
xlabel={Heading Error (deg)},
xmin=0,xmax=3,ymin=30,ymax=72,
xtick={0,0.5,1,1.5,2,2.5,3},
xticklabels={0,0.5,1,1.5,2,2.5,3},
]
\addplot+[nf] coordinates {(0,42.45) (0.5,42.45) (1,42.45) (1.5,42.45) (2,42.45) (2.5,42.45) (3,42.45)};
\addplot+[fc] coordinates {(0,49.83) (0.5,44.06) (1,38.52) (1.5,36.24) (2,34.82) (2.5,33.46) (3,32.15)};
\addplot+[dn] coordinates {(0,51.31) (0.5,46.84) (1,41.26) (1.5,38.74) (2,36.84) (2.5,35.12) (3,33.76)};
\addplot+[v2vn] coordinates {(0,61.71) (0.5,54.18) (1,43.63) (1.5,38.12) (2,35.46) (2.5,33.02) (3,31.48)};
\addplot+[v2xvit] coordinates {(0,58.37) (0.5,55.84) (1,45.59) (1.5,40.28) (2,37.12) (2.5,34.87) (3,33.10)};
\addplot+[when] coordinates {(0,54.96) (0.5,52.68) (1,45.02) (1.5,39.84) (2,36.06) (2.5,33.84) (3,32.24)};
\addplot+[where] coordinates {(0,67.68) (0.5,61.92) (1,48.90) (1.5,39.86) (2,36.42) (2.5,33.74) (3,31.63)};
\addplot+[how] coordinates {(0,68.76) (0.5,64.42) (1,56.24) (1.5,52.16) (2,49.03) (2.5,46.12) (3,43.70)};
\addplot+[cosdh] coordinates {(0,68.91) (0.5,64.18) (1,50.96) (1.5,46.72) (2,43.26) (2.5,40.56) (3,38.28)};
\addplot+[cobev] coordinates {(0,66.91) (0.5,61.18) (1,48.32) (1.5,39.54) (2,36.10) (2.5,33.48) (3,31.27)};
\addplot+[coalign] coordinates {(0,68.72) (0.5,65.34) (1,59.02) (1.5,57.36) (2,55.84) (2.5,54.62) (3,53.48)};
\addplot+[codyn] coordinates {(0,67.97) (0.5,62.54) (1,53.57) (1.5,50.42) (2,47.64) (2.5,45.18) (3,43.12)};
\addplot+[ours] coordinates {(0,69.42) (0.5,66.38) (1,63.08) (1.5,61.74) (2,60.62) (2.5,59.28) (3,58.10)};

\nextgroupplot[
xlabel={Packet Loss Ratio (\%)},
ylabel={AP@0.7},
xmin=0,xmax=90,ymin=15,ymax=56,
xtick={0,10,30,50,70,90},
]
\addplot+[nf] coordinates {(0,30.78) (10,30.78) (30,30.78) (50,30.78) (70,30.78) (90,30.78)};
\addplot+[fc] coordinates {(0,34.21) (10,33.15) (30,31.38) (50,29.63) (70,28.06) (90,25.82)};
\addplot+[dn] coordinates {(0,35.77) (10,34.62) (30,33.41) (50,32.27) (70,30.46) (90,28.84)};
\addplot+[v2vn] coordinates {(0,46.25) (10,43.82) (30,39.72) (50,34.49) (70,29.46) (90,24.38)};
\addplot+[v2xvit] coordinates {(0,42.77) (10,40.91) (30,38.31) (50,35.68) (70,33.29) (90,30.91)};
\addplot+[when] coordinates {(0,36.15) (10,35.08) (30,32.98) (50,31.40) (70,29.67) (90,27.24)};
\addplot+[where] coordinates {(0,52.18) (10,50.64) (30,46.42) (50,41.80) (70,36.04) (90,31.72)};
\addplot+[how] coordinates {(0,53.08) (10,51.42) (30,47.68) (50,42.94) (70,38.56) (90,34.12)};
\addplot+[cosdh] coordinates {(0,53.14) (10,51.76) (30,48.02) (50,43.28) (70,39.12) (90,35.46)};
\addplot+[cobev] coordinates {(0,51.64) (10,50.28) (30,45.96) (50,41.36) (70,35.82) (90,31.42)};
\addplot+[coalign] coordinates {(0,53.06) (10,50.86) (30,47.02) (50,42.06) (70,37.58) (90,33.72)};
\addplot+[codyn] coordinates {(0,52.36) (10,51.02) (30,48.16) (50,43.58) (70,39.84) (90,36.62)};
\addplot+[ours] coordinates {(0,53.78) (10,53.06) (30,51.84) (50,50.77) (70,48.63) (90,45.96)};

\nextgroupplot[
xlabel={Transmission Delay (ms)},
xmin=0,xmax=500,ymin=15,ymax=56,
xtick={0,100,200,300,400,500},
]
\addplot+[nf] coordinates {(0,30.78) (100,30.78) (200,30.78) (300,30.78) (400,30.78) (500,30.78)};
\addplot+[fc] coordinates {(0,34.21) (100,29.37) (200,27.84) (300,26.35) (400,25.22) (500,24.46)};
\addplot+[dn] coordinates {(0,35.77) (100,31.55) (200,30.21) (300,29.26) (400,28.34) (500,27.88)};
\addplot+[v2vn] coordinates {(0,46.25) (100,34.83) (200,30.68) (300,29.18) (400,28.06) (500,27.34)};
\addplot+[v2xvit] coordinates {(0,42.77) (100,34.04) (200,30.72) (300,29.28) (400,28.54) (500,27.96)};
\addplot+[when] coordinates {(0,36.15) (100,33.11) (200,30.84) (300,29.52) (400,28.72) (500,28.05)};
\addplot+[where] coordinates {(0,52.18) (100,41.42) (200,37.18) (300,35.84) (400,34.96) (500,34.61)};
\addplot+[how] coordinates {(0,53.08) (100,48.13) (200,45.92) (300,43.98) (400,42.36) (500,41.12)};
\addplot+[cosdh] coordinates {(0,53.14) (100,42.07) (200,37.84) (300,35.06) (400,33.82) (500,32.91)};
\addplot+[cobev] coordinates {(0,51.64) (100,42.36) (200,37.92) (300,36.42) (400,35.48) (500,35.06)};
\addplot+[coalign] coordinates {(0,53.06) (100,39.84) (200,35.62) (300,33.74) (400,32.61) (500,31.74)};
\addplot+[codyn] coordinates {(0,52.36) (100,47.56) (200,44.98) (300,43.18) (400,41.92) (500,40.78)};
\addplot+[ours] coordinates {(0,53.78) (100,47.94) (200,45.26) (300,43.62) (400,42.48) (500,41.53)};

\nextgroupplot[
xlabel={Heading Error (deg)},
xmin=0,xmax=3,ymin=15,ymax=56,
xtick={0,0.5,1,1.5,2,2.5,3},
xticklabels={0,0.5,1,1.5,2,2.5,3},
]
\addplot+[nf] coordinates {(0,30.78) (0.5,30.78) (1,30.78) (1.5,30.78) (2,30.78) (2.5,30.78) (3,30.78)};
\addplot+[fc] coordinates {(0,34.21) (0.5,30.96) (1,27.13) (1.5,25.81) (2,24.67) (2.5,23.82) (3,23.06)};
\addplot+[dn] coordinates {(0,35.77) (0.5,33.24) (1,30.94) (1.5,29.32) (2,28.26) (2.5,27.41) (3,26.88)};
\addplot+[v2vn] coordinates {(0,46.25) (0.5,37.52) (1,27.28) (1.5,23.82) (2,21.40) (2.5,19.82) (3,18.52)};
\addplot+[v2xvit] coordinates {(0,42.77) (0.5,40.21) (1,30.36) (1.5,27.48) (2,25.26) (2.5,23.62) (3,22.54)};
\addplot+[when] coordinates {(0,36.15) (0.5,34.36) (1,29.30) (1.5,26.74) (2,24.82) (2.5,23.12) (3,21.94)};
\addplot+[where] coordinates {(0,52.18) (0.5,44.08) (1,31.51) (1.5,26.34) (2,23.46) (2.5,21.40) (3,20.08)};
\addplot+[how] coordinates {(0,53.08) (0.5,49.16) (1,42.16) (1.5,38.72) (2,36.04) (2.5,33.82) (3,31.96)};
\addplot+[cosdh] coordinates {(0,53.14) (0.5,48.56) (1,35.94) (1.5,32.58) (2,29.84) (2.5,27.76) (3,26.18)};
\addplot+[cobev] coordinates {(0,51.64) (0.5,43.78) (1,31.18) (1.5,26.12) (2,23.24) (2.5,21.18) (3,19.94)};
\addplot+[coalign] coordinates {(0,53.06) (0.5,50.24) (1,43.67) (1.5,42.06) (2,40.68) (2.5,39.57) (3,38.82)};
\addplot+[codyn] coordinates {(0,52.36) (0.5,47.36) (1,38.56) (1.5,36.20) (2,34.06) (2.5,32.37) (3,31.04)};
\addplot+[ours] coordinates {(0,53.78) (0.5,51.52) (1,45.39) (1.5,44.08) (2,42.91) (2.5,41.80) (3,40.74)};
\end{groupplot}
\end{tikzpicture}}
\vspace{-0.18cm}
\makebox[\textwidth][c]{\resizebox{\textwidth}{!}{\pgfplotslegendfromname{robustlegend}}}
\vspace{-0.25cm}
\caption{\textbf{Robustness evaluation under varying packet loss ratios, transmission delays, and spatial misalignments.} The clean point and representative disturbance levels are aligned with the DAIR-V2X robustness comparison in Table~\ref{tab:robustness_dair}.}
\label{fig:robustness_different_conditions}
\vspace{-0.2cm}
\end{figure*}
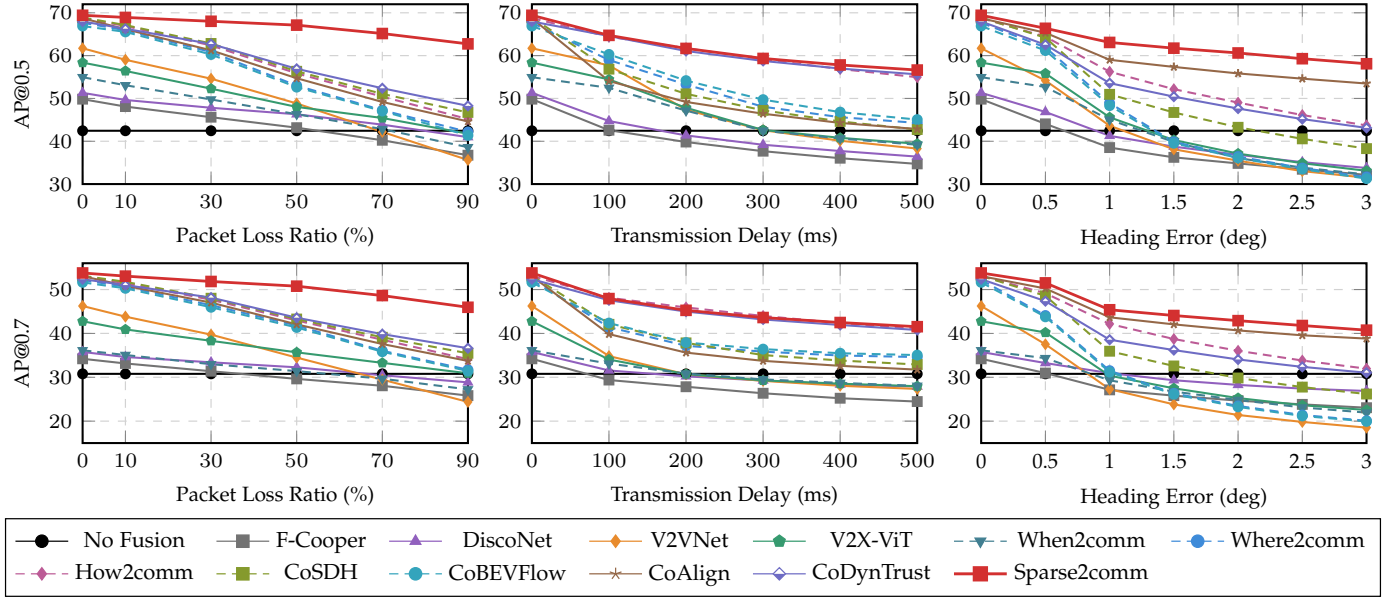

%% file: latex/tab/exps_ablation_modules.tex
\begin{comment}
\begin{table}[h]
\setlength{\abovecaptionskip}{6pt}
\addtolength{\tabcolsep}{0.0pt}
\caption{\textbf{Ablation study of the core components.} FE, LAA, and SCF denote Sparse Feature Encoding, Latency-Aware Alignment, and Self-Calibrating Fusion, respectively.}
\label{tab:module_ablation}
\scriptsize
\centering
\begin{tabularx}{1.0\linewidth}{c|c|c|cc|cc}
\toprule
\multirow{2}{*}{\textbf{SFE}} & \multirow{2}{*}{\textbf{LAA}} & \multirow{2}{*}{\textbf{SCF}}
& \multicolumn{2}{c|}{\textbf{Ideal}} & \multicolumn{2}{c}{\textbf{Noisy}} \\
\cmidrule(r){4-7}
& & & \textbf{AP@0.5$\uparrow$} & \textbf{AP@0.7$\uparrow$} & \textbf{AP@0.5$\uparrow$} & \textbf{AP@0.7$\uparrow$} \\
\midrule
~ & ~ & ~ & 67.68 & 52.18 & 37.95 & 27.75 \\
\checkmark & ~ & ~ & 66.31 & 51.12 & 49.91 & 36.52 \\
~ & \checkmark & ~ & 68.12 & 52.87 & 41.67 & 29.90 \\
~ & ~ & \checkmark & 68.76 & 53.24 & 48.53 & 33.32 \\
\checkmark & \checkmark & ~ & 67.93 & 52.54 & 52.82 & 35.78 \\
~ & \checkmark & \checkmark & \textbf{69.77} & \textbf{53.96} & 51.73 & 34.53 \\
\checkmark & ~ & \checkmark & 68.64 & 52.80 & \underline{55.68} & \underline{37.41} \\
\checkmark & \checkmark & \checkmark & \underline{69.42} & \underline{53.78} & \textbf{58.10} & \textbf{39.54} \\
\bottomrule
\end{tabularx}
\vspace{-0.2cm}
\end{table}
\end{comment}

\begin{table}[h]
\setlength{\abovecaptionskip}{6pt}
\addtolength{\tabcolsep}{-2.2pt}
\caption{\textbf{Ablation study of the core components.} 
SFE, LAA, and SCF denote Sparse Feature Encoding, Latency-Aware Alignment, and Self-Calibrating Fusion, respectively.}
\label{tab:module_ablation}
\scriptsize
\centering
\begin{tabularx}{1.0\linewidth}{c|c|c|cc|cc|c}
\toprule
\multirow{2}{*}{\textbf{SFE}} 
& \multirow{2}{*}{\textbf{LAA}} 
& \multirow{2}{*}{\textbf{SCF}} 
& \multicolumn{2}{c|}{\textbf{Ideal}} 
& \multicolumn{2}{c|}{\textbf{Noisy}} 
& \multirow{2}{*}{\textbf{BW$\downarrow$}} \\
\cmidrule(r){4-7}
& & & \textbf{AP@0.5$\uparrow$} & \textbf{AP@0.7$\uparrow$} 
& \textbf{AP@0.5$\uparrow$} & \textbf{AP@0.7$\uparrow$} &  (Mbps)\\
\midrule
~ & ~ & ~ & 67.68 & 52.18 & 37.95 & 27.75 & 12.9 \\
\checkmark & ~ & ~ & 66.31 & 51.12 & 49.91 & 36.52 & \textbf{2.6} \\
~ & \checkmark & ~ & 68.12 & 52.87 & 41.67 & 29.90 & 12.9 \\
~ & ~ & \checkmark & 68.76 & 53.24 & 48.53 & 33.32 & 12.9 \\
\checkmark & \checkmark & ~ & 67.93 & 52.54 & 52.82 & 35.78 & \textbf{2.6} \\
~ & \checkmark & \checkmark & \textbf{69.77} & \textbf{53.96} & 51.73 & 34.53 & 12.9 \\
\checkmark & ~ & \checkmark & 68.64 & 52.80 & \underline{55.68} &  \underline{37.41} & \textbf{2.6} \\
\checkmark & \checkmark & \checkmark & \underline{69.42} & \underline{53.78} & \textbf{58.10} & \textbf{39.54} & \textbf{2.6} \\
\bottomrule
\end{tabularx}
\vspace{-0.2cm}
\end{table}

%% file: latex/tab/exps_ablation_multi_stage_training.tex
\begin{table}[h]
\setlength{\abovecaptionskip}{6pt}
\addtolength{\tabcolsep}{0.5pt}
\caption{\textbf{Ablation study of the training strategy.}} 
\label{tab:training_ablation}
\scriptsize
\centering
\begin{tabularx}{1.0\linewidth}{c|cc|cc}
\toprule
\multirow{2}{*}{\textbf{Training Strategy}} 
& \multicolumn{2}{c|}{\textbf{Ideal}} 
& \multicolumn{2}{c}{\textbf{Noisy}} \\
\cmidrule(r){2-5}
& \textbf{AP@0.5$\uparrow$} & \textbf{AP@0.7$\uparrow$} & \textbf{AP@0.5$\uparrow$} & \textbf{AP@0.7$\uparrow$} \\
\midrule
Joint Training & 58.65 & 48.23 & 49.40  & 33.77   \\
Multi-Stage Training &  \textbf{69.42} & \textbf{53.78}  &  \textbf{58.10} &  \textbf{39.54} \\
\bottomrule
\end{tabularx}
\vspace{-0.2cm}
\end{table}

%% file: latex/tab/exps_ablation_sampling.tex
\begin{table}[h!t]
\setlength{\abovecaptionskip}{8pt}
\addtolength{\tabcolsep}{0pt}
\caption{\textbf{Ablation study on masking strategies in the Sparse Feature Encoding module.}} 
\label{tab:masking_ablation}
\scriptsize
\centering
\begin{tabularx}{1.0\linewidth}{c|cc|cc}
\toprule
\multirow{2}{*}{\textbf{Sample Strategy}} 
& \multicolumn{2}{c|}{\textbf{Ideal}} 
& \multicolumn{2}{c}{\textbf{Noisy}} \\
\cmidrule(r){2-5}
& \textbf{AP@0.5$\uparrow$} & \textbf{AP@0.7$\uparrow$} & \textbf{AP@0.5$\uparrow$} & \textbf{AP@0.7$\uparrow$} \\
\midrule
Uniform Sampling & 67.25 & 52.30 & 52.97 & 36.10 \\
Random Sampling & 68.76 & 53.12 & 56.23 & 38.52 \\
Object-aware Sampling & \textbf{69.42} & \textbf{53.78} & \textbf{58.10} & \textbf{39.54} \\
\bottomrule
\end{tabularx}
\vspace{-0.2cm}
\end{table}

%% file: latex/tab/exp_ablation_data_augmentation.tex
\begin{table}[h!t]
\setlength{\abovecaptionskip}{6pt}
\setlength{\tabcolsep}{2.0pt}
\renewcommand{\arraystretch}{1.2}
\caption{\textbf{Ablation study on the data augmentation.}
RM denotes random masking on collaborator BEV features, with mask ratio sampled from $r \sim \mathcal{U}(0, 1.0)$ (Uniform distribution). PD denotes pose disturbance applied to BEV features, including rotational noise $\Delta R \sim \mathcal{N}(0, 3.0^\circ)$ and translational noise $\Delta T \sim \mathcal{N}(0, 0.5)$m.(Gaussian distribution).}
\label{tab:data_aug_ablation}
\scriptsize
\centering
\begin{tabular}{l|cc|cc|cc}
\toprule
\multirow{2}{*}{\textbf{Method}} 
& \multicolumn{2}{c|}{\textbf{Data augmentation}} 
& \multicolumn{2}{c|}{\textbf{Ideal}} 
& \multicolumn{2}{c}{\textbf{Noisy}} \\
\cmidrule(r){2-7}
& RM & PD 
& AP@0.5$\uparrow$ & AP@0.7$\uparrow$ 
& AP@0.5$\uparrow$ & AP@0.7$\uparrow$ \\
\midrule
\multirow{4}{*}{Baseline}  &  &  & \underline{67.68} & \underline{52.18} & 37.95 & 27.75 \\
~ & \checkmark &  & 66.41 & 49.92 & 40.66 & 26.57 \\
~ &  & \checkmark & 64.56 & 47.58 & 42.96 & \underline{29.98} \\
% ~ & \checkmark & \checkmark & 38.92 & 28.66 & \underline{49.46} & \underline{32.12} \\
~ & \checkmark & \checkmark & 67.03 & 49.86 & \underline{45.47} & 28.43 \\
\midrule
\textbf{Sparse2comm} 
& \checkmark & \checkmark 
& \textbf{69.42}
& \textbf{53.78} 
& \textbf{58.1} 
& \textbf{39.54} \\
\bottomrule
\end{tabular}
\vspace{-0.2cm}
\end{table}

%% file: latex/tab/exp_sota_efficiency.tex
\begin{table*}[t]
\setlength{\abovecaptionskip}{8pt}
\renewcommand{\arraystretch}{1.12}
\setlength{\tabcolsep}{1.6pt}
\setlength{\aboverulesep}{0pt}
\setlength{\belowrulesep}{0pt}
\caption{\textbf{End-to-end efficiency comparison of the representative methods.}
All methods are profiled under the same DAIR-V2X input setting. Model latency denotes pure inference latency, while transmission time is computed from the transmitted bandwidth under conservative 10 Mbps and optimistic 50 Mbps V2X link rates. Gray entries denote the single-agent No Fusion baseline, and black bold numbers indicate the best cooperative-method result in each column.}
\label{tab:efficiency_comparison}
\scriptsize
\centering
\begin{tabularx}{\textwidth}{@{}l l *{6}{>{\centering\arraybackslash}X} >{\columncolor{black!6}\centering\arraybackslash}X@{}>{\columncolor{black!6}\centering\arraybackslash}X@{}}
\toprule
\multirow{3}{*}{\textbf{Method}}
& \multirow{3}{*}{\textbf{Category}}
& \multicolumn{3}{c}{\textbf{Computation}}
& \multicolumn{3}{c}{\textbf{Communication}}
& \multicolumn{2}{>{\columncolor{black!6}}c@{}}{\textbf{End-to-End Latency}} \\
\cmidrule(lr){3-5}\cmidrule(lr){6-8}\cmidrule(lr){9-10}
&
& \textbf{Params}
& \textbf{FLOPs}
& \textbf{Model Lat.}
& \textbf{BW}
& \textbf{Tx.@10M}
& \textbf{Tx.@50M}
& \textbf{E2E@10M}
& \textbf{E2E@50M} \\
&
& \textbf{(M)$\downarrow$}
& \textbf{(G)$\downarrow$}
& \textbf{(ms)$\downarrow$}
& \textbf{(Mbps)$\downarrow$}
& \textbf{(ms)$\downarrow$}
& \textbf{(ms)$\downarrow$}
& \textbf{(ms)$\downarrow$}
& \textbf{(ms)$\downarrow$} \\
\midrule
\multicolumn{8}{l}{\textit{Basic and classic cooperative perception}} & \multicolumn{2}{>{\columncolor{black!6}}c@{}}{} \\
\textcolor{gray}{No Fusion} & \textcolor{gray}{Single-agent} & \textcolor{gray}{6.6} & \textcolor{gray}{118.4} & \textcolor{gray}{37.8} & \textcolor{gray}{0} & \textcolor{gray}{0} & \textcolor{gray}{0} & \textcolor{gray}{37.8} & \textcolor{gray}{37.8} \\
F-Cooper~\cite{chen2019f} & Full feature & 7.3 & 225.6 & \textbf{48.9} & 258.1 & 2581.0 & 516.2 & 2629.9 & 565.1 \\
DiscoNet~\cite{mehr2019disconet} & Full feature & \textbf{6.7} & 238.2 & 50.8 & 258.1 & 2581.0 & 516.2 & 2631.8 & 567.0 \\
V2VNet~\cite{wang2020v2vnet} & Full feature & 14.6 & 531.6 & 88.5 & 258.1 & 2581.0 & 516.2 & 2669.5 & 604.7 \\
V2X-ViT~\cite{xu2022v2x} & Full feature & 14.2 & 287.0 & 85.5 & 258.1 & 2581.0 & 516.2 & 2666.5 & 601.7 \\
\midrule
\multicolumn{8}{l}{\textit{Communication-efficient cooperative perception}} & \multicolumn{2}{>{\columncolor{black!6}}c@{}}{} \\
When2comm~\cite{liu2020when2com} & Adaptive comm. & 8.8 & 228.3 & 52.7 & 258.1 & 2581.0 & 516.2 & 2633.7 & 568.9 \\
Where2comm~\cite{hu2022where2comm} & Selective feature & 11.4 & \textbf{218.9} & 51.2 & 12.9 & 129.0 & 25.8 & 180.2 & 77.0 \\
How2comm~\cite{yang2023how2comm} & Pragmatic comm. & 16.2 & 283.6 & 64.8 & 8.8 & 88.0 & 17.6 & 152.8 & 82.4 \\
CoSDH~\cite{xu2025cosdh} & Hybrid fusion & 17.9 & 296.8 & 67.2 & 7.2 & 72.0 & 14.4 & 139.2 & 81.6 \\
\midrule
\multicolumn{8}{l}{\textit{Robust cooperative perception}} & \multicolumn{2}{>{\columncolor{black!6}}c@{}}{} \\
CoBEVFlow~\cite{wei2023asynchrony} & BEV flow alignment & 18.9 & 301.7 & 66.4 & 14.5 & 145.0 & 29.0 & 211.4 & 95.4 \\
CoAlign~\cite{10160546} & Pose alignment & 18.4 & 334.2 & 73.5 & 258.1 & 2581.0 & 516.2 & 2654.5 & 589.7 \\
CoDynTrust~\cite{xu2025codyntrust} & Async. robustness & 27.6 & 382.5 & 82.9 & 27.4 & 274.0 & 54.8 & 356.9 & 137.7 \\
\midrule
\textbf{Sparse2comm} & Progressive restoration & 34.0 & 326.4 & 69.7 & \textbf{2.6} & \textbf{26.0} & \textbf{5.2} & \textbf{95.7} & \textbf{74.9} \\
\bottomrule
\end{tabularx}
\vspace{-0.25cm}
\end{table*}

%% file: latex/tab/exp_ablation_efficiency.tex
\begin{table}[h]
\setlength{\abovecaptionskip}{6pt}
\addtolength{\tabcolsep}{0.8pt}
\caption{\textbf{Ablation study of Sparse2comm efficiency.}
SFE, LAA, and SCF denote Sparse Feature Encoding, Latency-Aware Alignment, and Self-calibrating Fusion. We report inference-stage parameters, FLOPs, and latency under the same profiling setting.}
\label{tab:efficiency_ablation}
\scriptsize
\centering
\begin{tabularx}{1.0\linewidth}{c|c|c|ccc}
\toprule
\textbf{SFE} & \textbf{LAA} & \textbf{SCF} 
& \textbf{Params (M)} & \textbf{FLOPs (G)} & \textbf{Latency (ms)} \\
\midrule
~ & ~ & ~ & 11.4 & 218.9 & 51.2 \\
\checkmark & ~ & ~ & 24.0 & 269.3 & 59.8 \\
~ & \checkmark & ~ & 16.0 & 255.7 & 56.2 \\
~ & ~ & \checkmark & 16.8 & 239.2 & 54.0 \\
\checkmark & \checkmark & ~ & 28.6 & 306.1 & 64.7 \\
~ & \checkmark & \checkmark & 21.4 & 276.0 & 59.0 \\
\checkmark & ~ & \checkmark & 29.4 & 289.6 & 62.5 \\
\checkmark & \checkmark & \checkmark & 34.0 & 326.4 & 69.7 \\
\bottomrule
\end{tabularx}
\vspace{-0.4cm}
\end{table}

%% file: latex/tab/exp_ablation_sfe_config.tex
\begin{table}[h!t]
\setlength{\abovecaptionskip}{6pt}
\addtolength{\tabcolsep}{-2.8pt}
\caption{\textbf{Ablation study on the architecture of Sparse Feature Encoding.}
$N_E$ and $N_D$ denote the number of ViT blocks in the encoder and decoder, respectively.}
\label{tab:exp_ablation_sfe_config}
\scriptsize
\centering
\begin{tabularx}{1.0\linewidth}{l|cc|cc|cc}
\toprule
\multirow{2}{*}{\textbf{Setting}} 
& \multicolumn{2}{c|}{\textbf{\#Blocks}} 
& \multicolumn{2}{c|}{\textbf{Ideal}} 
& \multicolumn{2}{c}{\textbf{Noisy}} \\
\cmidrule(r){2-3} \cmidrule(r){4-7}
& $N_E$ & $N_D$
& AP@0.5$\uparrow$ & AP@0.7$\uparrow$
& AP@0.5$\uparrow$ & AP@0.7$\uparrow$ \\
\midrule
S1: Shallow          & 3 & 1  & 66.23 & 51.12 & 53.17 & 35.14 \\
S2: Encoder-heavy    & 7 & 1  & 68.47 & 53.18 & 56.41 & 38.23 \\
S3: Balanced (ours)  & 5 & 2  & \textbf{69.42} & \textbf{53.78} & \textbf{58.10} & \textbf{39.54} \\
S4: Decoder-heavy    & 5 & 4  & 69.27 & 53.63 & 57.86 & 39.31 \\
S5: Deep (optional)  & 7 & 4  & 69.14 & 53.46 & 57.22 & 38.87 \\
\bottomrule
\end{tabularx}
\vspace{-0.2cm}
\end{table}

%% file: latex/fig/visualization.tex
\begin{figure*}[!t]
\centering
\setlength{\abovecaptionskip}{3pt}
\includegraphics[width=1.0\textwidth]{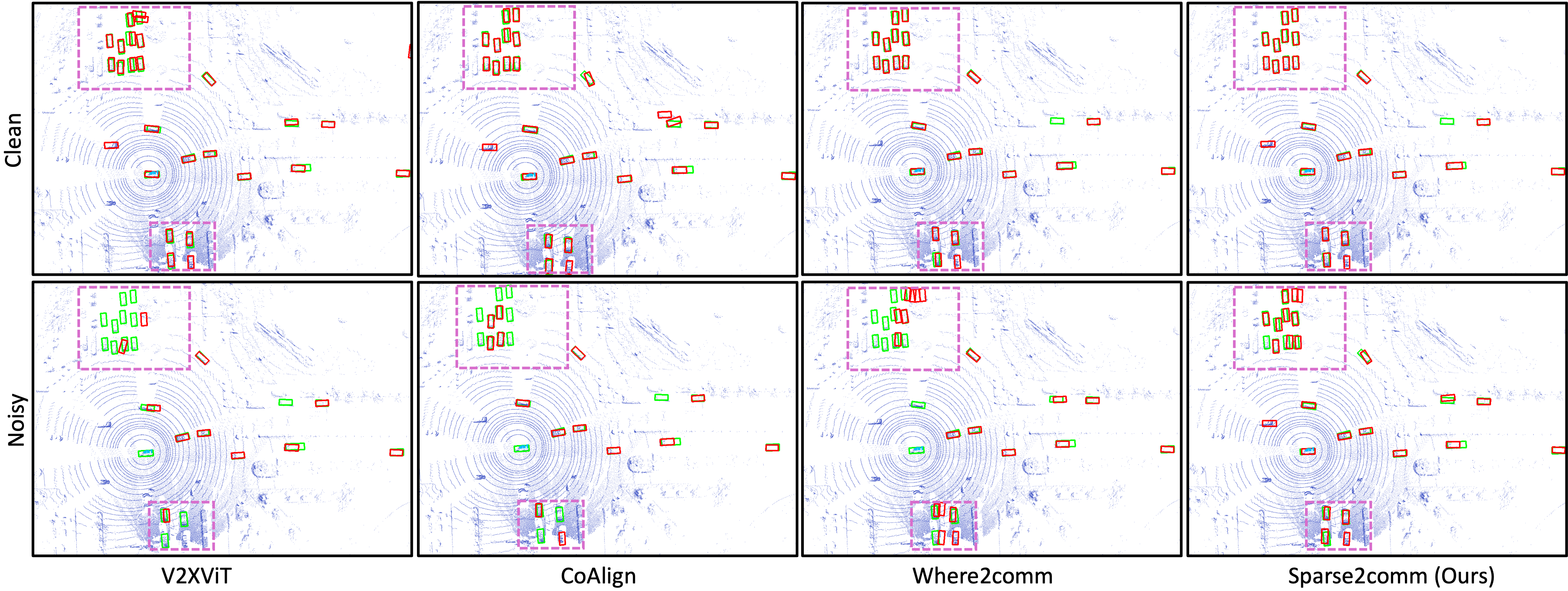}
\caption{\textbf{Detection result visualization on DAIR-V2X.} Representative examples are shown under clean and mixed noisy settings. \textcolor{green}{Green} and \textcolor{red}{red} boxes denote the ground truths and detection results, respectively. 
Under packet loss, transmission delay, and spatial misalignment, existing methods exhibit noticeable false and missed detections, while our \textbf{Sparse2comm} maintains accurate and consistent bounding boxes.
}
\label{fig:qualitative_results}
\vspace{-0.2cm}
\end{figure*}

\begin{figure*}[!t]
\centering
\setlength{\abovecaptionskip}{3pt}
\includegraphics[width=1.0\textwidth]{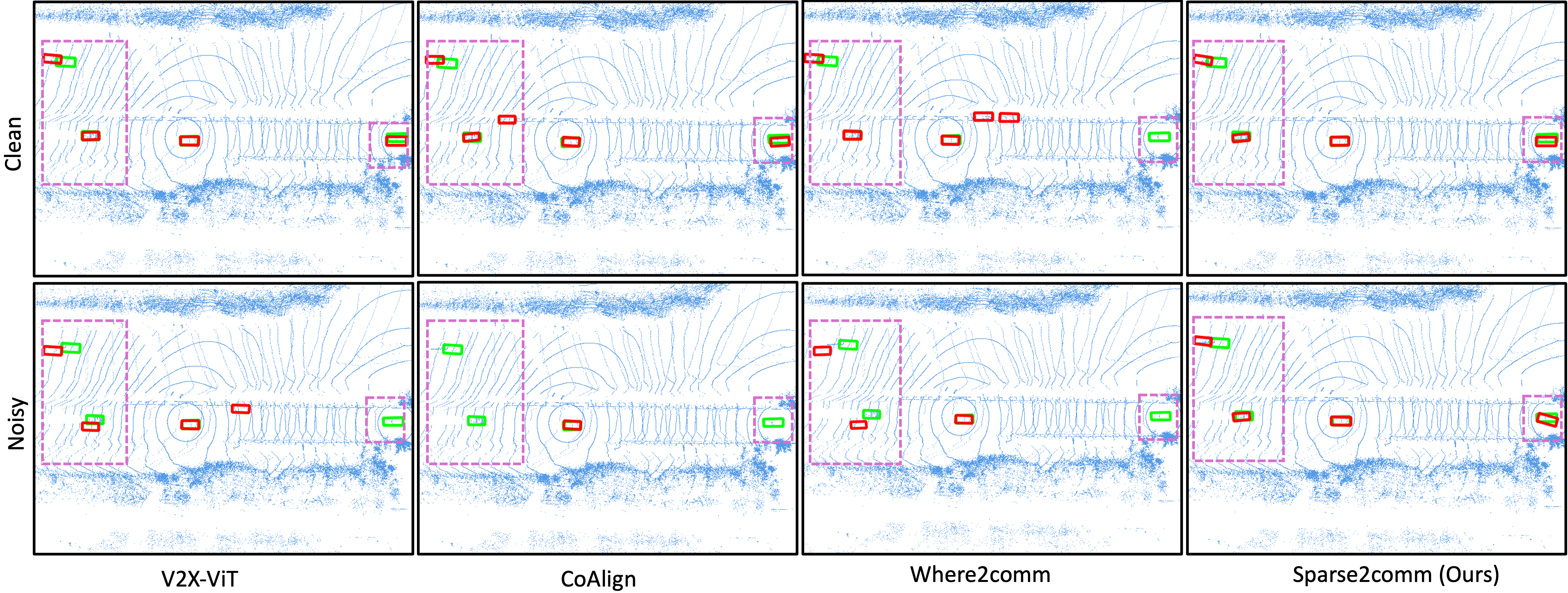}
\caption{\textbf{Detection result visualization on V2V4Real.} Additional real-world vehicle--vehicle examples are shown under clean and mixed noisy settings. \textcolor{green}{Green} and \textcolor{red}{red} boxes denote the ground truths and detection results, respectively. Sparse2comm preserves more spatially consistent detections when cooperative features are affected by packet loss, transmission delay, and spatial misalignment.}
\label{fig:qualitative_v2v4real}
\vspace{-0.2cm}
\end{figure*}

%% file: latex/fig/sfe_module_visualization.tex
\begin{figure*}[h!t]
\centering
\setlength{\abovecaptionskip}{-5pt}
\includegraphics[width=1.0\textwidth]{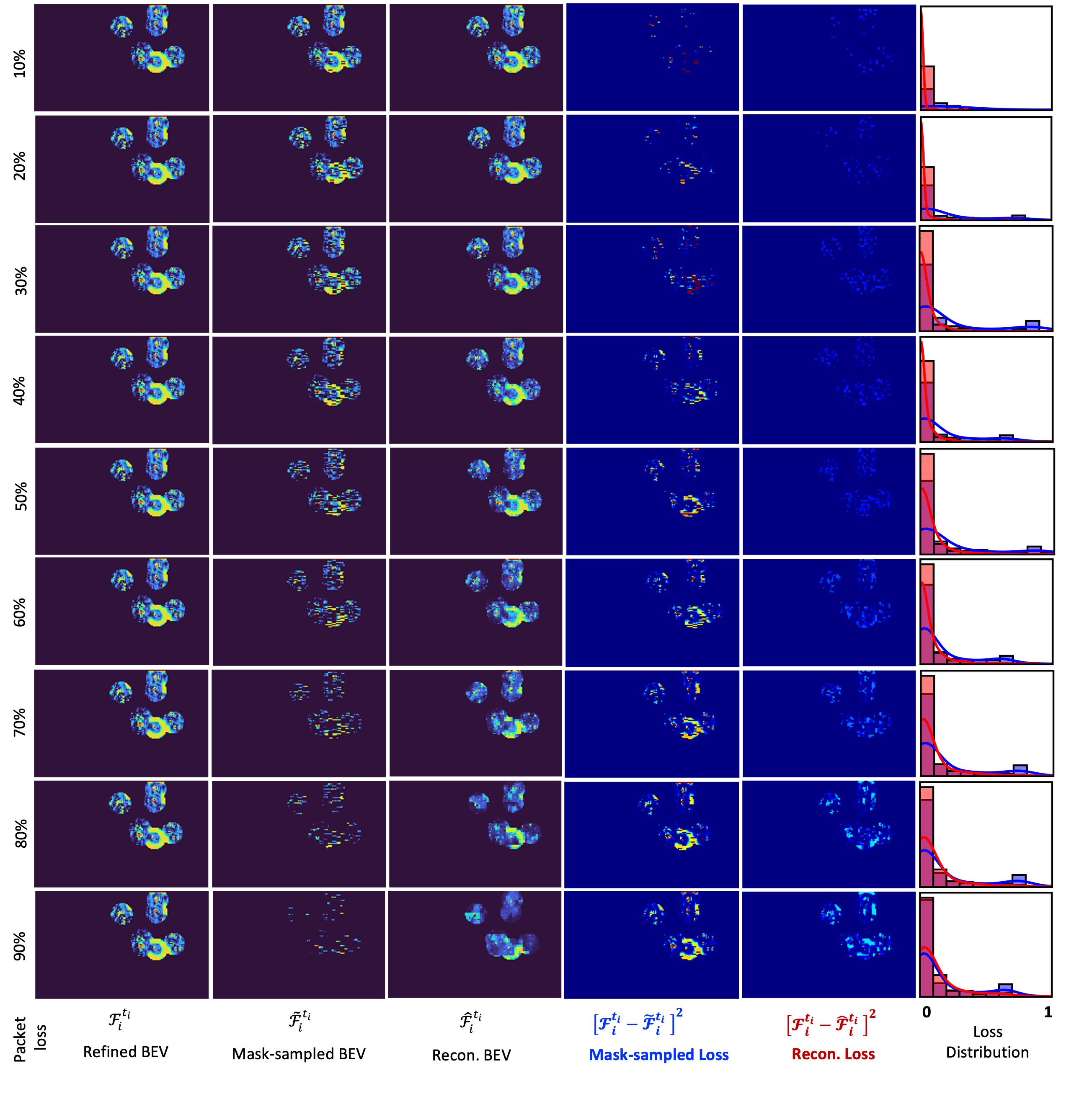}
\caption{\textbf{Visualization of the Sparse Feature Encoding (SFE).} We visualize the refined BEV features, the mask-sampled BEV features obtained by the Mask Sampling submodule, the reconstructed BEV features produced by the Feature Reconstruction submodule, along with their corresponding information-loss maps and loss histograms.
% We show refined BEV features, mask-sampled BEV features that simulate packet loss, reconstructed BEV features, and their corresponding information-loss maps and histograms. 
The mask-sampled BEV features suffer from severe information degradation, while the reconstruction submodule effectively recovers the missing content, producing features close to the refined BEV and loss values near zero. 
This behavior remains consistent across packet-loss ratios from 10\% to 100\%, demonstrating SFE’s strong capability to recover missing information and reduce bandwidth through sparse-to-dense reconstruction.}
\label{fig:sfe_visualization}
\vspace{-0.2cm}
\end{figure*}

%% file: latex/sec/5_conclusion.tex
\section{Conclusion}
This paper presented Sparse2comm, a bandwidth-efficient and robust cooperative 3D object detection framework for unreliable real-world cooperation. Rather than treating communication reduction, packet loss, temporal delay, and spatial misalignment as separate problems, Sparse2comm formulates them as coupled degradations of cooperative features and addresses them through a progressive feature restoration pipeline. Sparse Feature Encoding builds a compact and recoverable object-centric representation for ultra-low-bandwidth transmission and packet-loss recovery. Latency-Aware Alignment then predicts motion flow from temporal collaborator features to compensate delayed cooperative messages at the ego-vehicle timestamp, while Self-Calibrating Fusion estimates residual pose-induced offsets in a self-supervised manner to refine cross-agent feature correspondence before adaptive fusion.
Experiments on DAIR-V2X~\cite{yu2022dair}, OpenV2V~\cite{xu2022opv2v}, and V2V4Real~\cite{xu2023v2v4real} show that Sparse2comm maintains competitive clean accuracy and achieves stronger mixed robustness across diverse cooperative perception scenarios. Compared with the representative selective communication baseline Where2comm, Sparse2comm improves mixed-setting AP@0.5/AP@0.7 by \textbf{+20.15/+11.79}, \textbf{+12.66/+11.07}, and \textbf{+15.36/+12.61} on the three datasets, respectively, while using only \textbf{1.0\%} relative feature bandwidth.

\noindent \textbf{Limitation and future work.}
Sparse2comm currently focuses on cooperative 3D object detection with feature-level communication. Future work will extend the progressive restoration idea to broader cooperative autonomy tasks, such as occupancy perception, trajectory prediction, and end-to-end cooperative driving, and further investigate its deployment behavior under real communication systems with dynamic bandwidth and latency.

%% file: main.bib
@String(ICLR = {Int. Conf. Learn. Represent.})

@String(AAAI = {AAAI})

@String(ICLR  = {ICLR})

@inproceedings{wang2020v2vnet,
  title={V2vnet: Vehicle-to-vehicle communication for joint perception and prediction},
  author={Wang, Tsun-Hsuan and Manivasagam, Sivabalan and Liang, Ming and Yang, Bin and Zeng, Wenyuan and Urtasun, Raquel},
  booktitle={European conference on computer vision},
  pages={605--621},
  year={2020},
  organization={Springer}
}

@inproceedings{yu2022dair,
  title={Dair-v2x: A large-scale dataset for vehicle-infrastructure cooperative 3d object detection},
  author={Yu, Haibao and Luo, Yizhen and Shu, Mao and Huo, Yiyi and Yang, Zebang and Shi, Yifeng and Guo, Zhenglong and Li, Hanyu and Hu, Xing and Yuan, Jirui and others},
  booktitle={Proceedings of the IEEE/CVF conference on computer vision and pattern recognition},
  pages={21361--21370},
  year={2022}
}

@inproceedings{yu2025end,
  title={End-to-end autonomous driving through v2x cooperation},
  author={Yu, Haibao and Yang, Wenxian and Zhong, Jiaru and Yang, Zhenwei and Fan, Siqi and Luo, Ping and Nie, Zaiqing},
  booktitle={Proceedings of the AAAI Conference on Artificial Intelligence},
  volume={39},
  number={9},
  pages={9598--9606},
  year={2025}
}

@inproceedings{yu2023v2x,
  title={V2x-seq: A large-scale sequential dataset for vehicle-infrastructure cooperative perception and forecasting},
  author={Yu, Haibao and Yang, Wenxian and Ruan, Hongzhi and Yang, Zhenwei and Tang, Yingjuan and Gao, Xu and Hao, Xin and Shi, Yifeng and Pan, Yifeng and Sun, Ning and others},
  booktitle={Proceedings of the IEEE/CVF Conference on Computer Vision and Pattern Recognition},
  pages={5486--5495},
  year={2023}
}

@article{yu2023flow,
  title={Flow-based feature fusion for vehicle-infrastructure cooperative 3d object detection},
  author={Yu, Haibao and Tang, Yingjuan and Xie, Enze and Mao, Jilei and Luo, Ping and Nie, Zaiqing},
  journal={Advances in Neural Information Processing Systems},
  volume={36},
  pages={34493--34503},
  year={2023}
}

@article{li2022v2xsim,
  title={{V2X-Sim}: Multi-agent collaborative perception dataset and benchmark for autonomous driving},
  author={Li, Yiming and Ma, Dekun and An, Ziyan and Wang, Zixun and Zhong, Yiqi and Chen, Siheng and Feng, Chen},
  journal={IEEE Robotics and Automation Letters},
  volume={7},
  number={4},
  pages={10914--10921},
  year={2022},
  doi={10.1109/LRA.2022.3192802}
}

@inproceedings{xu2022opv2v,
  title={{OPV2V}: An open benchmark dataset and fusion pipeline for perception with vehicle-to-vehicle communication},
  author={Xu, Runsheng and Xiang, Hao and Xia, Xin and Han, Xu and Li, Jinlong and Ma, Jiaqi},
  booktitle={2022 IEEE International Conference on Robotics and Automation (ICRA)},
  pages={2583--2589},
  year={2022},
  organization={IEEE},
  doi={10.1109/ICRA46639.2022.9812038}
}

@inproceedings{xu2023cobevt,
  title={{CoBEVT}: Cooperative bird's eye view semantic segmentation with sparse transformers},
  author={Xu, Runsheng and Tu, Zhengzhong and Xiang, Hao and Shao, Wei and Zhou, Bolei and Ma, Jiaqi},
  booktitle={Proceedings of The 6th Conference on Robot Learning},
  pages={989--1000},
  year={2023},
  volume={205},
  series={Proceedings of Machine Learning Research},
  publisher={PMLR}
}

@article{gao2024survey,
  title={A survey of collaborative perception in intelligent vehicles at intersections},
  author={Gao, Xin and Zhang, Xinyu and Lu, Yiguo and Huang, Yuning and Yang, Lei and Xiong, Yijin and Liu, Peng},
  journal={IEEE Transactions on Intelligent Vehicles},
  year={2024},
  publisher={IEEE}
}

@article{hu2022where2comm,
  title={Where2comm: Communication-efficient collaborative perception via spatial confidence maps},
  author={Hu, Yue and Fang, Shaoheng and Lei, Zixing and Zhong, Yiqi and Chen, Siheng},
  journal={Advances in neural information processing systems},
  volume={35},
  pages={4874--4886},
  year={2022}
}

@article{liu2024v2x,
  title={V2x-pc: Vehicle-to-everything collaborative perception via point cluster},
  author={Liu, Si and Ding, Zihan and Fu, Jiahui and Li, Hongyu and Chen, Siheng and Zhang, Shifeng and Zhou, Xu},
  journal={arXiv preprint arXiv:2403.16635},
  year={2024}
}

@article{ren2024interruption,
  title={Interruption-aware cooperative perception for V2X communication-aided autonomous driving},
  author={Ren, Shunli and Lei, Zixing and Wang, Zi and Dianati, Mehrdad and Wang, Yafei and Chen, Siheng and Zhang, Wenjun},
  journal={IEEE Transactions on Intelligent Vehicles},
  volume={9},
  number={4},
  pages={4698--4714},
  year={2024},
  publisher={IEEE}
}

@inproceedings{hong2024mrcnet,
  title={Multi-agent Collaborative Perception via Motion-aware Robust Communication Network},
  author={Hong, Shixin and Liu, Yu and Li, Zhi and Li, Shaohui and He, You},
  booktitle={Proceedings of the IEEE/CVF Conference on Computer Vision and Pattern Recognition},
  pages={15301--15310},
  year={2024}
}

@inproceedings{xu2025codyntrust,
  title={CoDynTrust: Robust Asynchronous Collaborative Perception via Dynamic Feature Trust Modulus},
  author={Xu, Yunjiang and Li, Lingzhi and Wang, Jin and Yang, Benyuan and Wu, Zhiwen and Chen, Xinhong and Wang, Jianping},
  booktitle={2025 IEEE International Conference on Robotics and Automation (ICRA)},
  pages={336--342},
  year={2025},
  doi={10.1109/ICRA55743.2025.11127779}
}

@INPROCEEDINGS{10160546,
  author={Lu, Yifan and Li, Quanhao and Liu, Baoan and Dianati, Mehrdad and Feng, Chen and Chen, Siheng and Wang, Yanfeng},
  booktitle={2023 IEEE International Conference on Robotics and Automation (ICRA)}, 
  title={Robust Collaborative 3D Object Detection in Presence of Pose Errors}, 
  year={2023},
  volume={},
  number={},
  pages={4812-4818},
  doi={10.1109/ICRA48891.2023.10160546}}

@article{clancy2024wireless,
  title={Wireless access for V2X communications: Research, challenges and opportunities},
  author={Clancy, Joseph and Mullins, Darragh and Deegan, Brian and Horgan, Jonathan and Ward, Enda and Eising, Ciar{\'a}n and Denny, Patrick and Jones, Edward and Glavin, Martin},
  journal={IEEE Communications Surveys \& Tutorials},
  volume={26},
  number={3},
  pages={2082--2119},
  year={2024},
  publisher={IEEE}
}

@article{yang2023how2comm,
  title={How2comm: Communication-efficient and collaboration-pragmatic multi-agent perception},
  author={Yang, Dingkang and Yang, Kun and Wang, Yuzheng and Liu, Jing and Xu, Zhi and Yin, Rongbin and Zhai, Peng and Zhang, Lihua},
  journal={Advances in Neural Information Processing Systems},
  volume={36},
  pages={25151--25164},
  year={2023}
}

@inproceedings{tang2025rocooper,
  title={RoCooper: Robust Cooperative Perception Under Vehicle-to-Vehicle Communication Impairments},
  author={Tang, Tao and Zhang, Chaokun and Chen, Gong and Xia, Feng and Xu, Qimei and Liu, Changle},
  booktitle={IEEE INFOCOM 2025-IEEE Conference on Computer Communications},
  pages={1--10},
  year={2025},
  organization={IEEE}
}

@inproceedings{xu2022v2x,
  title={V2x-vit: Vehicle-to-everything cooperative perception with vision transformer},
  author={Xu, Runsheng and Xiang, Hao and Tu, Zhengzhong and Xia, Xin and Yang, Ming-Hsuan and Ma, Jiaqi},
  booktitle={European conference on computer vision},
  pages={107--124},
  year={2022},
  organization={Springer}
}

@inproceedings{song2023graphalign,
  title={Graphalign: Enhancing accurate feature alignment by graph matching for multi-modal 3d object detection},
  author={Song, Ziying and Wei, Haiyue and Bai, Lin and Yang, Lei and Jia, Caiyan},
  booktitle={Proceedings of the IEEE/CVF international conference on computer vision},
  pages={3358--3369},
  year={2023}
}

@inproceedings{song2023cooperative,
  title={A cooperative perception system robust to localization errors},
  author={Song, Zhiying and Wen, Fuxi and Zhang, Hailiang and Li, Jun},
  booktitle={2023 IEEE Intelligent Vehicles Symposium (IV)},
  pages={1--6},
  year={2023},
  organization={IEEE}
}

@article{li2023learning,
  title={Learning for vehicle-to-vehicle cooperative perception under lossy communication},
  author={Li, Jinlong and Xu, Runsheng and Liu, Xinyu and Ma, Jin and Chi, Zicheng and Ma, Jiaqi and Yu, Hongkai},
  journal={IEEE Transactions on Intelligent Vehicles},
  volume={8},
  number={4},
  pages={2650--2660},
  year={2023},
  publisher={IEEE}
}

@inproceedings{liu2020when2com,
  title={When2com: Multi-agent perception via communication graph grouping},
  author={Liu, Yen-Cheng and Tian, Junjiao and Glaser, Nathaniel and Kira, Zsolt},
  booktitle={Proceedings of the IEEE/CVF Conference on computer vision and pattern recognition},
  pages={4106--4115},
  year={2020}
}

@inproceedings{song2025traf,
  title={Traf-align: Trajectory-aware feature alignment for asynchronous multi-agent perception},
  author={Song, Zhiying and Yang, Lei and Wen, Fuxi and Li, Jun},
  booktitle={Proceedings of the Computer Vision and Pattern Recognition Conference},
  pages={12048--12057},
  year={2025}
}

@inproceedings{lei2022latency,
  title={Latency-aware collaborative perception},
  author={Lei, Zixing and Ren, Shunli and Hu, Yue and Zhang, Wenjun and Chen, Siheng},
  booktitle={European Conference on Computer Vision},
  pages={316--332},
  year={2022},
  organization={Springer}
}

@article{wei2023asynchrony,
  title={Asynchrony-robust collaborative perception via bird's eye view flow},
  author={Wei, Sizhe and Wei, Yuxi and Hu, Yue and Lu, Yifan and Zhong, Yiqi and Chen, Siheng and Zhang, Ya},
  journal={Advances in Neural Information Processing Systems},
  volume={36},
  pages={28462--28477},
  year={2023}
}

@inproceedings{ding2025point,
  title={Point cluster: A compact message unit for communication-efficient collaborative perception},
  author={Ding, Zihan and Fu, Jiahui and Liu, Si and Li, Hongyu and Chen, Siheng and Li, Hongsheng and Zhang, Shifeng and Zhou, Xu},
  booktitle={The Thirteenth International Conference on Learning Representations},
  year={2025}
}

@article{yang2024v2x,
  title={{V2X-Radar}: A multi-modal dataset with {4D} radar for cooperative perception},
  author={Yang, Lei and Zhang, Xinyu and Li, Jun and Wang, Chen and Ma, Jiaqi and Song, Zhiying and Zhao, Tong and Song, Ziying and Wang, Li and Zhou, Mo and Shen, Yang and Wu, Kai and Lv, Chen},
  journal={Advances in Neural Information Processing Systems},
  year={2025}
}

@inproceedings{xu2023v2v4real,
  title={V2v4real: A real-world large-scale dataset for vehicle-to-vehicle cooperative perception},
  author={Xu, Runsheng and Xia, Xin and Li, Jinlong and Li, Hanzhao and Zhang, Shuo and Tu, Zhengzhong and Meng, Zonglin and Xiang, Hao and Dong, Xiaoyu and Song, Rui and others},
  booktitle={Proceedings of the IEEE/CVF conference on computer vision and pattern recognition},
  pages={13712--13722},
  year={2023}
}

@inproceedings{xiang2024v2x,
  title={V2X-Real: a large-scale dataset for vehicle-to-everything cooperative perception},
  author={Xiang, Hao and Zheng, Zhaoliang and Xia, Xin and Xu, Runsheng and Gao, Letian and Zhou, Zewei and Han, Xu and Ji, Xinkai and Li, Mingxi and Meng, Zonglin and others},
  booktitle={European Conference on Computer Vision},
  pages={455--470},
  year={2024},
  organization={Springer}
}

@inproceedings{wang2021fcos3d,
  title={Fcos3d: Fully convolutional one-stage monocular 3d object detection},
  author={Wang, Tai and Zhu, Xinge and Pang, Jiangmiao and Lin, Dahua},
  booktitle={Proceedings of the IEEE/CVF international conference on computer vision},
  pages={913--922},
  year={2021}
}

@article{li2024bevformer,
  title={Bevformer: learning bird's-eye-view representation from lidar-camera via spatiotemporal transformers},
  author={Li, Zhiqi and Wang, Wenhai and Li, Hongyang and Xie, Enze and Sima, Chonghao and Lu, Tong and Yu, Qiao and Dai, Jifeng},
  journal={IEEE Transactions on Pattern Analysis and Machine Intelligence},
  year={2024},
  publisher={IEEE}
}

@inproceedings{li2023bevdepth,
  title={Bevdepth: Acquisition of reliable depth for multi-view 3d object detection},
  author={Li, Yinhao and Ge, Zheng and Yu, Guanyi and Yang, Jinrong and Wang, Zengran and Shi, Yukang and Sun, Jianjian and Li, Zeming},
  booktitle={Proceedings of the AAAI conference on artificial intelligence},
  volume={37},
  number={2},
  pages={1477--1485},
  year={2023}
}

@inproceedings{zhou2018voxelnet,
  title={Voxelnet: End-to-end learning for point cloud based 3d object detection},
  author={Zhou, Yin and Tuzel, Oncel},
  booktitle={Proceedings of the IEEE conference on computer vision and pattern recognition},
  pages={4490--4499},
  year={2018}
}

@inproceedings{lang2019pointpillars,
  title={Pointpillars: Fast encoders for object detection from point clouds},
  author={Lang, Alex H and Vora, Sourabh and Caesar, Holger and Zhou, Lubing and Yang, Jiong and Beijbom, Oscar},
  booktitle={Proceedings of the IEEE/CVF conference on computer vision and pattern recognition},
  pages={12697--12705},
  year={2019}
}

@article{yan2018second,
  title={Second: Sparsely embedded convolutional detection},
  author={Yan, Yan and Mao, Yuxing and Li, Bo},
  journal={Sensors},
  volume={18},
  number={10},
  pages={3337},
  year={2018},
  publisher={Multidisciplinary Digital Publishing Institute}
}

@article{liu2022bevfusion,
  title={Bevfusion: Multi-task multi-sensor fusion with unified bird's-eye view representation},
  author={Liu, Zhijian and Tang, Haotian and Amini, Alexander and Yang, Xinyu and Mao, Huizi and Rus, Daniela and Han, Song},
  journal={arXiv preprint arXiv:2205.13542},
  year={2022}
}

@inproceedings{song2024graphbev,
  title={Graphbev: Towards robust bev feature alignment for multi-modal 3d object detection},
  author={Song, Ziying and Yang, Lei and Xu, Shaoqing and Liu, Lin and Xu, Dongyang and Jia, Caiyan and Jia, Feiyang and Wang, Li},
  booktitle={European Conference on Computer Vision},
  pages={347--366},
  year={2024},
  organization={Springer}
}

@inproceedings{yang2023bevheight,
  title={Bevheight: A robust framework for vision-based roadside 3d object detection},
  author={Yang, Lei and Yu, Kaicheng and Tang, Tao and Li, Jun and Yuan, Kun and Wang, Li and Zhang, Xinyu and Chen, Peng},
  booktitle={Proceedings of the IEEE/CVF Conference on Computer Vision and Pattern Recognition},
  pages={21611--21620},
  year={2023}
}

@article{yang2025bevheight++,
  title={Bevheight++: Toward robust visual centric 3d object detection},
  author={Yang, Lei and Tang, Tao and Li, Jun and Chen, Peng and Yuan, Kun and Wang, Li and Huang, Yi and Zhang, Xinyu and Yu, Kaicheng},
  journal={IEEE Transactions on Pattern Analysis and Machine Intelligence},
  year={2025},
  publisher={IEEE}
}

@article{yang2023lite,
  title={Lite-fpn for keypoint-based monocular 3d object detection},
  author={Yang, Lei and Zhang, Xinyu and Li, Jun and Wang, Li and Zhu, Minghan and Zhu, Lei},
  journal={Knowledge-Based Systems},
  volume={271},
  pages={110517},
  year={2023},
  publisher={Elsevier}
}

@article{yang2023mix,
  title={Mix-teaching: A simple, unified and effective semi-supervised learning framework for monocular 3d object detection},
  author={Yang, Lei and Zhang, Xinyu and Li, Jun and Wang, Li and Zhu, Minghan and Zhang, Chuang and Liu, Huaping},
  journal={IEEE Transactions on Circuits and Systems for Video Technology},
  volume={33},
  number={11},
  pages={6832--6844},
  year={2023},
  publisher={IEEE}
}

@inproceedings{he2016deep,
  title={Deep residual learning for image recognition},
  author={He, Kaiming and Zhang, Xiangyu and Ren, Shaoqing and Sun, Jian},
  booktitle={Proceedings of the IEEE conference on computer vision and pattern recognition},
  pages={770--778},
  year={2016}
}

@article{vaswani2017attention,
  title={Attention is all you need},
  author={Vaswani, Ashish and Shazeer, Noam and Parmar, Niki and Uszkoreit, Jakob and Jones, Llion and Gomez, Aidan N and Kaiser, {\L}ukasz and Polosukhin, Illia},
  journal={Advances in neural information processing systems},
  volume={30},
  year={2017}
}

@inproceedings{ioffe2015batch,
  title={Batch normalization: Accelerating deep network training by reducing internal covariate shift},
  author={Ioffe, Sergey and Szegedy, Christian},
  booktitle={International conference on machine learning},
  pages={448--456},
  year={2015},
  organization={pmlr}
}

@inproceedings{nair2010rectified,
  title={Rectified linear units improve restricted boltzmann machines},
  author={Nair, Vinod and Hinton, Geoffrey E},
  booktitle={Proceedings of the 27th international conference on machine learning (ICML-10)},
  pages={807--814},
  year={2010}
}

@article{kingma2014adam,
  title={Adam: A method for stochastic optimization},
  author={Kingma, Diederik P and Ba, Jimmy},
  journal={arXiv preprint arXiv:1412.6980},
  year={2014}
}

@inproceedings{mehr2019disconet,
  title={Learning Distilled Collaboration Graph for Multi-Agent Perception},
  author={Li, Yiming and Ren, Shunli and Wu, Pengxiang and Chen, Siheng and Feng, Chen and Zhang, Wenjun},
  booktitle={Advances in Neural Information Processing Systems},
  volume={34},
  year={2021}
}

@inproceedings{chen2019f,
  title={F-cooper: Feature based cooperative perception for autonomous vehicle edge computing system using 3D point clouds},
  author={Chen, Qi and Ma, Xu and Tang, Sihai and Guo, Jingda and Yang, Qing and Fu, Song},
  booktitle={Proceedings of the 4th ACM/IEEE Symposium on Edge Computing},
  pages={88--100},
  year={2019}
}

@inproceedings{yang2023what2comm,
  title={What2comm: Towards communication-efficient collaborative perception via feature decoupling},
  author={Yang, Kun and Yang, Dingkang and Zhang, Jingyu and Wang, Hanqi and Sun, Peng and Song, Liang},
  booktitle={Proceedings of the 31st ACM international conference on multimedia},
  pages={7686--7695},
  year={2023}
}

@inproceedings{zhang2024ermvp,
  title={Ermvp: Communication-efficient and collaboration-robust multi-vehicle perception in challenging environments},
  author={Zhang, Jingyu and Yang, Kun and Wang, Yilei and Wang, Hanqi and Sun, Peng and Song, Liang},
  booktitle={Proceedings of the IEEE/CVF Conference on Computer Vision and Pattern Recognition},
  pages={12575--12584},
  year={2024}
}

@inproceedings{xu2025cosdh,
  title={CoSDH: Communication-Efficient Collaborative Perception via Supply-Demand Awareness and Intermediate-Late Hybridization},
  author={Xu, Junhao and Zhang, Yanan and Cai, Zhi and Huang, Di},
  booktitle={Proceedings of the Computer Vision and Pattern Recognition Conference},
  pages={6834--6843},
  year={2025}
}

@article{dosovitskiy2020vit,
  title={An Image is Worth 16x16 Words: Transformers for Image Recognition at Scale},
  author={Dosovitskiy, Alexey and Beyer, Lucas and Kolesnikov, Alexander and Weissenborn, Dirk and Zhai, Xiaohua and Unterthiner, Thomas and  Dehghani, Mostafa and Minderer, Matthias and Heigold, Georg and Gelly, Sylvain and Uszkoreit, Jakob and Houlsby, Neil},
  journal={ICLR},
  year={2021}
}

@article{liu2025toward,
  author = {Liu, Genjia and Hu, Yue and Xu, Chenxin and Mao, Weibo and Ge, Junhao and Huang, Zhengxiang and Lu, Yifan and Xu, Yinda and Xia, Junkai and Wang, Yafei and Chen, Siheng},
  title = {Toward Collaborative Autonomous Driving: Simulation Platform and End-to-End System},
  journal = {IEEE Transactions on Pattern Analysis and Machine Intelligence},
  year = {2025},
  volume = {47},
  number = {8},
  pages = {6566--6584},
  doi = {10.1109/TPAMI.2025.3560327}
}

@article{xu2025v2xvitv2,
  author = {Xu, Runsheng and Chen, Chia-Ju and Tu, Zhengzhong and Yang, Ming-Hsuan},
  title = {{V2X-ViTv2}: Improved Vision Transformers for Vehicle-to-Everything Cooperative Perception},
  journal = {IEEE Transactions on Pattern Analysis and Machine Intelligence},
  year = {2025},
  volume = {47},
  number = {1},
  pages = {650--662},
  doi = {10.1109/TPAMI.2024.3479222}
}

@article{hu2026pragmatic,
  author = {Hu, Yue and Pang, Xianghe and Qin, Xiaoqi and Eldar, Yonina C. and Chen, Siheng and Zhang, Ping and Zhang, Wenjun},
  title = {Pragmatic Communication in Multi-Agent Collaborative Perception},
  journal = {IEEE Transactions on Pattern Analysis and Machine Intelligence},
  year = {2026},
  volume = {48},
  number = {8},
  pages = {9279--9296},
  doi = {10.1109/TPAMI.2026.3680062}
}
